\documentclass[runningheads]{llncs}

\usepackage{eccv}

\usepackage{eccvabbrv}

\usepackage{graphicx}
\usepackage{booktabs}

\usepackage{color}

\newcommand{\model}{\textsc{MetaCap}}
\newcommand{\dataset}{\textsc{WildDynaCap}}

\usepackage{graphicx}
\usepackage{amsfonts}

\usepackage{wrapfig,lipsum,booktabs}
\usepackage{algorithmicx}
\usepackage{algorithm}
\usepackage{algpseudocode}

\usepackage{booktabs}
\usepackage{multirow}
\usepackage{tabularx}
\usepackage{soul}
 
\usepackage{ulem}

\definecolor{bronze}{rgb}{1,1,0.6}
\definecolor{silve}{rgb}{0.969,0.796,0.600}
\definecolor{gold}{rgb}{0.941,0.592,0.600}

\newcommand{\gold}[1]{\colorbox{gold}{{#1}}}
\newcommand{\silve}[1]{\colorbox{silve}{{#1}}}
\newcommand{\bronze}[1]{\colorbox{bronze}{{#1}}}

\usepackage[accsupp]{axessibility}  

\usepackage[colorlinks]{hyperref}

\usepackage{orcidlink}

\begin{document}

\title{\model: Meta-learning Priors from Multi-View Imagery for Sparse-view Human Performance Capture and Rendering}

\titlerunning{\model{}}

\author{Guoxing Sun\inst{1},
        Rishabh Dabral\inst{1},
        Pascal Fua\inst{2},
        Christian Theobalt\inst{1}, \\
        Marc Habermann\inst{1}
}

\authorrunning{Sun et al.}

\institute{Max Planck Institute for Informatics, Saarland Informatics Campus \and
EPFL \\
\email{\{gsun,rdabral,theobalt,mhaberma\}@mpi-inf.mpg.de, pascal.fua@epfl.ch}\\
\url{https://vcai.mpi-inf.mpg.de/projects/MetaCap/}
}

\maketitle

%
%
%
%
\begin{figure}
  \centering
  \vspace{-22pt}
  \includegraphics[width=0.95\linewidth]{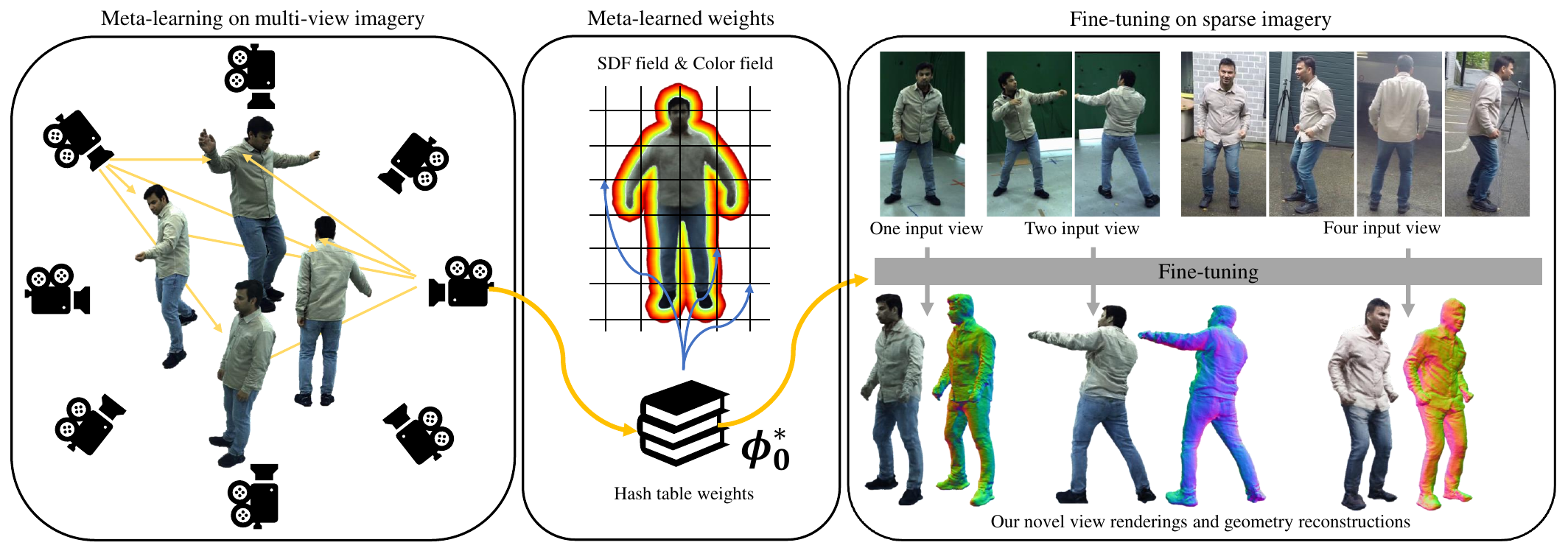}
  \vspace{-8pt}
  
  \caption{
        We propose \model, a new approach for capturing 3D humans from sparse-view or even monocular images, which enables high fidelity 3D geometry recovery and photoreal free-view synthesis within a range of seconds to minutes.
        }
    \label{fig:futureTeas}
\end{figure}
\vspace{-30pt}

%
%
\begin{abstract} \label{sec:abstract}
Faithful human performance capture and free-view rendering from sparse RGB observations is a long-standing problem in Vision and Graphics. 
The main challenges are the lack of observations and the inherent ambiguities of the setting, e.g. occlusions and depth ambiguity. 
As a result, radiance fields, which have shown great promise in capturing high-frequency appearance and geometry details in dense setups, perform poorly when na\"ively supervising them on sparse camera views, as the field simply overfits to the sparse-view inputs.
To address this, we propose \model, a method for efficient and high-quality geometry recovery and novel view synthesis given very sparse or even a single view of the human. 
Our key idea is to meta-learn the radiance field weights solely from potentially sparse multi-view videos, which can serve as a prior when fine-tuning them on sparse imagery depicting the human. 
This prior provides a good network weight initialization, thereby effectively addressing ambiguities in sparse-view capture.
Due to the articulated structure of the human body and motion-induced surface deformations, learning such a prior is non-trivial. 
Therefore, we propose to meta-learn the field weights in a pose-canonicalized space, which reduces the spatial feature range and makes feature learning more effective.
Consequently, one can fine-tune our field parameters to quickly generalize to unseen poses, novel illumination conditions as well as novel and sparse (even monocular) camera views.
For evaluating our method under different scenarios, we collect a new dataset, \dataset, which contains subjects captured in, both, a dense camera dome and in-the-wild sparse camera rigs, and demonstrate superior results compared to recent state-of-the-art methods on, both, public and \dataset{} dataset.
\vspace{-6pt}
\end{abstract}
%
%
%
%
\section{Introduction} \label{sec:introduction}
Human performance capture has made enormous strides in recent years. 
Typically, the quality of the reconstruction improves with the number of cameras being used. 
The highest-quality results are obtained using many---8 or more---calibrated cameras in a dome-like configuration~\cite{collet2015high, guo2019relightables,shao2022diffustereo, zhao2022human, shuai2022multinb}. 
However, such capture domes are expensive and difficult to set up, which restricts their applicability. 
A truly practical solution ought to be easy to deploy while supporting arbitrary camera configurations, including {\it sparse} ones, i.e. four or less cameras.
Unfortunately, sparsity comes with its own challenges, such as self-occlusions and depth ambiguities, which often result in lower quality outcomes.
%
%
\par
How to mitigate the effects of these ambiguities and missing information is the core problem of sparse-view reconstructions and researchers have explored different priors to compensate for the ambiguities of such an ill-posed problem.
Earlier works introduce priors on template meshes~\cite{deepcap,jiang2022hifecap,monoperfcap,livecap2019} to solve the above ambiguities. 
However, meshes suffer from limited resolution and are not easy to incorporate into learning frameworks.
Implicit surface representations offer a promising alternative and some methods~\cite{saito2019pifu,zheng2021deepmulticap} can learn implicit field priors from large-scale human scan datasets.
However, they suffer from blurry appearance and poor generalization capability because the amount of data is limited and model capacity is small.
While many works have recently focused on animatable implicit radiance fields for humans~\cite{weng_humannerf_2022_cvpr,liu2021neural,Zhao_2022_CVPR, peng2022animatable, ARAH:ECCV:2022,habermann2022hdhumans}, only few ~\cite{long2022sparseneus,wang2023sparsenerf, niemeyer2022regnerf, yang2022freenerf} have explored priors for optimizing or fitting a neural implicit field given sparse imagery.
They usually only show results for \textit{simple objects}, but not for complex structures like articulated humans.
Recently, some works~\cite{sitzmann2020metasdf, tancik2020meta} showed that meta-learning can learn powerful neural field priors for few-shot learning.
To learn the field prior for personalized high fidelity human capture more effectively, we argue that we should decompose the human model into a coarse-level geometry represented by mesh-based templates~\cite{SMPL:2015,habermann2021} and a fine-level neural implicit field~\cite{wang2021neus,wang2022neus2,zhao2022human} accounting for details such as clothing wrinkles.
The coarse-level prior from multi-view images is easier to obtain and can serve as a crucial component in fine-level prior learning.
At the fine level, we, for the first time in literature, {\it explore learning a meta prior from images for an implicit human representation}, which at test time can be optimized using sparse image cues, thereby allowing fast adaptation to unseen poses, novel views, different camera setups, and novel illumination conditions.  
%
%
\par
Our method, dubbed \model{}, represents the fine-level human body as a signed distance field (SDF) and an appearance field parameterized with an efficient hashgrid encoding~\cite{muller2022instant}, which can be volume rendered~\cite{wang2021neus} into an image.
To learn an effective prior for such a representation, we propose to meta-learn the hashgrid parameters solely from multi-view imagery during training in an end-to-end manner, i.e. learning the optimal weights, which, at test time under the sparse settings, yield improved convergence rate and accuracy.
However, na\"ively meta-learning these implicit parameters for static scenes~\cite{sitzmann2020metasdf, tancik2020meta}, leads to poor performance on complicated settings such as human performance capture (Fig.~\ref{fig:ablation}).
This is because the human surface is highly articulated and always deforms.
Thus, learning hashgrid parameters, which live on a spatially fixed grid, is not effective, e.g. the limbs can be mapped onto entirely different hash table parameters depending on their articulation.
Therefore, we further propose a space canonicalization step, i.e. we transform points from global space to a canonical pose space.
To obtain the transform, we query the nearest transformation of a coarse-level human template.
We highlight that this canonicalization is not tightly bound to a specific template, but supports deformable human models~\cite{habermann2021} as well as parametric (piece-wise rigid) body models~\cite{SMPL:2015}.
Additionally, we introduce occlusion handling, where we use a visibility map to guide the ray sampling, enabling reconstruction of heavily occluded regions at sparse or even monocular inputs at inference time.
%
%
\par
We evaluate our method on the DynaCap~\cite{habermann2021} benchmark, which provides multi-view recordings of humans in a controlled studio setup.
However, DynaCap lacks in-the-wild scenes for evaluating the robustness to various environment conditions such as the difference in lighting and camera parameters. 
Thus, we provide a new dataset, \dataset, containing paired \textit{dense multi-view studio} captures and \textit{sparse multi-view in-the-wild} recordings for each subject. 
In summary, our main contributions are:
\begin{itemize}
    \item A meta-learning method learned from multi-view imagery for high-quality reconstruction of human geometry and appearance under sparse cameras.
    \item At the technical core, we propose a meta-learning strategy to learn the optimal weights of an implicit human representation solely from multi-view images, which effectively serves as a prior when deployed to the sparse reconstruction task.
    \item We further demonstrate the importance of space canonicalization for the human-specific meta-learning task, and introduce a dedicated occlusion handling strategy.
\end{itemize} 
Our quantitative and qualitative experiments demonstrate that \model{} achieves state-of-the-art geometry recovery and novel view synthesis compared to prior works.
Our evaluations also demonstrate that \model{} generalizes to novel poses and induced surface deformations, change of lighting conditions, and camera parameters.
Moreover, we highlight the versatility of our approach as our meta-learning strategy can be supervised on an arbitrary number of views also including monocular videos.
Similarly, during fine-tuning, our approach supports reconstruction from sparse in-the-wild multi-view images as well as monocular imagery (see Fig.~\ref{fig:futureTeas}).
Last, our method is agnostic to the choice of the deformable human model, thus, even supporting loose types of apparel.
%
%
%
%
\section{Related Work} \label{sec:related}
In the following, we discuss prior works on (1) scene-agnostic implicit representations for novel view synthesis and reconstruction, (2) human-specific sparse-view reconstruction and rendering methods, (3) personalized performance capture and (4) 3D reconstruction using meta-learning.
Animatable avatar methods~\cite{CAPE:CVPR:20, Saito:CVPR:2021, habermann2021, liu2021neural,
zheng2022structured, li2022tava, habermann2022hdhumans, shen2023xavatar, kwon2023deliffas, Pang_2024_CVPR, relightneuralactor2024eccv, zhu2023trihuman}, which only take the skeletal pose as input during inference, are not in our scope, since we focus on the reconstruction task, i.e. recovering geometry and appearance from 2D imagery during inference.
%
%
\par 
\noindent\textbf{Scene-agnostic Implicit Representations.}\label{subsec:rw_scene}
The emergence of implicit representations for 3D scene reconstruction~\cite{mescheder2019occupancy, park2019deepsdf, saito2019pifu, wang2021neus} and novel-view synthesis~\cite{mildenhall2020nerf, barron2021mipnerf, barron2022mipnerf360, barron2023zipnerf} paved the way towards compact and high-fidelity scene representations.
Subsequent methods have attempted to improve upon a variety of aspects, such as faster training~\cite{muller2022instant, yu2022plenoxels, Chen2022ECCV}, faster inference~\cite{yu2021plenoctrees, kerbl3Dgaussians}, sparse~\cite{mvsnerf, long2022sparseneus, wang2023sparsenerf} and monocular-view~\cite{yu2021pixelnerf, gu2023nerfdiff} reconstruction.
In our work, we build upon the scene representation introduced in NeuS2~\cite{wang2022neus2} and Instant-NSR~\cite{zhao2022human}.
They leverage multi-grid hash encoding~\cite{muller2022instant} for parameterizing the implicit surface~\cite{wang2021neus} to enable fast and high quality reconstruction.
However, they require sufficiently dense multi-view images to ensure accurate geometry and appearance recovery.
Due to occlusions and the depth ambiguity inherently present in the sparse setup, simply optimizing such methods with sparse RGB imagery leads to poor reconstructions.
More recently, some methods explored different geometry priors~\cite{long2022sparseneus, wang2023sparsenerf} or regularizations~\cite{yang2022freenerf, niemeyer2022regnerf,johnson2023ub4d} to enable few-shot novel view synthesis. 
However, none of them proves to be effective under large camera baselines or sparse $360^{\circ}$ scenarios, where the image observations are very limited and occlusions happen more frequently.
We address these challenges by meta-learning an optimal set of initial hashgrid parameters, which effectively serves as a prior during inference on sparse observations, and a human-specific space canonicalization.
%
%
\par
\noindent\textbf{Sparse-View Performance Capture and Rendering.} \label{subsec:rw_human}
Beyond occlusions and the depth ambiguity, sparse performance capture comes with the additional challenges of modeling and capturing the body articulations as well as complex surface deformations of the clothes.
To compensate for the absence of dense supervision, sparse human reconstruction methods typically employ additional priors.
They range from using a parametric template for a coarse initialization of the geometry~\cite{zheng2021pamir, zheng2021deepmulticap, peng2021neural, peng2021animatable, peng2022animatable, shao2022floren}, to using depth supervision~\cite{yu2018doublefusion, zuo2020sparsefusion, su2020robustfusion, palafox2021spams, xue2023nsf} and data-driven priors~\cite{Zheng2019DeepHuman, saito2019pifu, saito2020pifuhd, zheng2021deepmulticap, MetaAvatar:NeurIPS:2021, shao2022diffustereo, xiu2022icon, xiu2023econ, huang2024tech,Davydov22a,TretschkNonRigidSurvey2023}.
Several volumetric rendering-based methods focus on learning the human geometry and appearance in the canonical space~\cite{peng2021neural, peng2021animatable, remelli2022drivable, shao2022doublefield, jiang2022neuman, wang2022nerfcap, Pan_2023_ICCV, Wang2023ClothedHumanCap}, which learns shared features across different poses. 
While such methods excel in novel view synthesis tasks, the recovered geometry is often coarse.
Some works~\cite{kwon2021neural, Mihajlovic:KeypointNeRF:ECCV2022, Pan_2023_ICCV, buhler2023preface} learn radiance field priors from large scale datasets.
However, their conventional learning strategy limits their fine-tuning ability.
Another key challenge in human reconstruction is associated with the modeling of loose clothing.
Previous methods have attempted to address this by optimizing the vertex deformations on top of a template model~\cite{CAPE:CVPR:20, xiang2020monoclothcap, livecap2019, deepcap, habermann2021} or by modeling clothes as a separate layer~\cite{de2023drapenet, Wang2023ClothedHumanCap}.
In this work, we utilize a hybrid representation where we leverage an explicit human model to facilitate meta-learning an optimal set of weights parameterizing an implicit field in a pose-canonicalized space. 
%
%
\par 
\noindent\textbf{Personalized Performance Capture.}\label{subsec:rw_person}
In contrast to generalizable performance capture methods~\cite{saito2019pifu,zheng2021deepmulticap,ARAH:ECCV:2022,Pan_2023_ICCV}, personalized human capture first builds a person-specific prior on multi-view data~\cite{deepcap, wang2022nerfcap, xiang2023drivable,Davydov22a,shetty2023holoported,li2022avatarcap, yueli2021} while during inference the avatar can be reconstructed from sparser signals, e.g. a monocular image. 
Therefore, the goal of such methods is less about cross-identity generalization, but rather highest-fidelity surface and appearance recovery from sparse sensory data while ensuring generalization to different human motions, lighting condition, camera poses, camera numbers, and surface dynamics.
Our setting is most closely related to these methods. However, we for the first time in literature propose meta-learning as a prior on the implicit human representation solely from images while experimentally showing that we outperform previous methods operating under the same setting.
%
%
\par 
\noindent\textbf{Reconstruction with Meta-Learning. }\label{subsec:rw_meta}
Meta-learning is intended to learn from multiple tasks or a single task~\cite{hospedales2021meta}. 
When new observations are presented, domain adaptation or improved performance can be achieved with minimal training iterations. 
Here, we focus on optimization-based methods~\cite{li2016learning, finn2017model, nichol2018first, antoniou2018train, rajeswaran2019meta}, especially MAML~\cite{finn2017model} and Reptile~\cite{nichol2018first}. 
MetaSDF~\cite{sitzmann2020metasdf} learns SDF priors and achieves faster inference than auto-decoder methods~\cite{park2019deepsdf}. 
Tancik et al.~\cite{tancik2020meta} apply a similar strategy to different coordinate-based neural representations, such as image fields and radiance fields~\cite{mildenhall2020nerf}.
MetaAvatar~\cite{MetaAvatar:NeurIPS:2021} extends this idea from static to dynamic human representations. 
After meta-learning on 3D human scans~\cite{CAPE:CVPR:20}, the learned SDF weights can be fine-tuned using a few depth maps.
\par 
ARAH~\cite{ARAH:ECCV:2022} is methodically most closely related to our work. 
However, there are significant differences:
1) They meta-learn human priors from 3D scans while only supporting image-only supervision during fine-tuning. 
In stark contrast, our formulation allows meta-learning from potentially sparse images directly as we deeply entangle the volume rendering and hash-grid parameterization with the meta-learning routine.
Our end-to-end prior learning and fine-tuning, thus, enables better recovery of high-frequency geometry and appearance.
2) Our whole design, i.e. space canonicalization and scene parameterization, is geared towards efficiency, significantly improving the fine-tuning time, reducing from hours (ARAH) to minutes (\model{}).
3) We experimentally show that our canonicalization is agnostic to specific template choices, thus, we can also account for loose clothing while ARAH solely shows tightly clothed people.
%
%
%
\begin{figure}[tb]
	\includegraphics[width=0.98\linewidth]{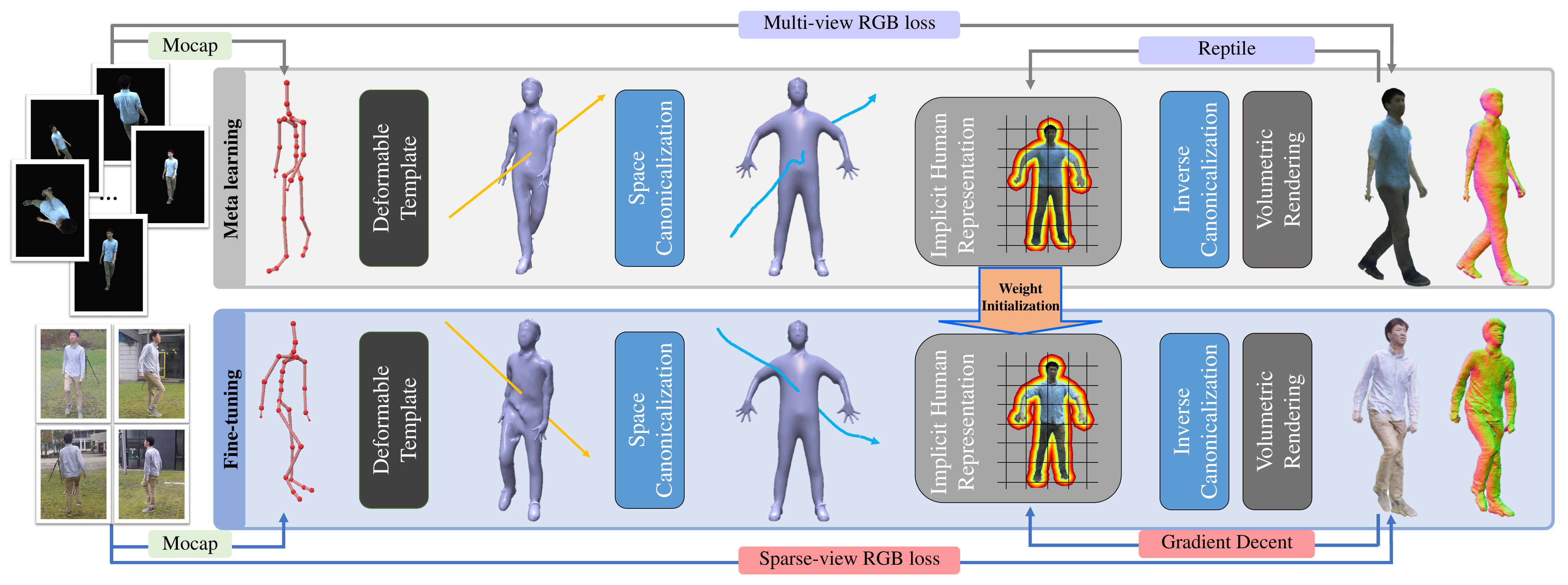}
 \vspace{-10pt}
	\caption
	{
	    \textbf{Overview}.
        \model{} is a novel performance capture method, which meta-learns the pose-canonical and optimal network weights of an implicit human representation solely from multi-view images.
        At inference, with only sparse in-the-wild images, those weights effectively act as a prior and resolve the inherent ambiguities, i.e. occlusion and depth, while maintaining high geometric and visual quality.
	}
        \vspace{-10pt}
	\label{fig:overview}
\end{figure}
%
%
%
%

%
%
\vspace{-10pt}
\section{Method} \label{sec:method}
Our goal is to recover high-quality human models from few, or even only one, RGB image.
Let us consider an implicit function $f_\phi$ parameterized by a network with weights $\phi$ that encodes the geometry and the appearance of the clothed human.
Our sparse-view reconstruction method comprises two phases, multi-view meta-learning followed by sparse-view fine-tuning, as shown in Fig.~\ref{fig:overview}.
\par 
During the meta-learning phase (Sec.~\ref{subsec:me_bigpicture} and \ref{subsec:me_vr}), we are given a multi-view video using $K_d$ cameras, which captures the performer.
We denote the images of frame $f$ as $\mathcal{I}_{f} = \{I_{f}^{k} | k \in [1;K_d]\}$ and assume the availability of paired masks $\mathcal{O}_{f}$ and the corresponding skeletal motion $\mathcal{M}_{f, W} = \{\boldsymbol{\theta}_{f-W}, \boldsymbol{\alpha}_{f-W}, \boldsymbol{z}_{f-W}, ..., \boldsymbol{\theta}_{f}, \boldsymbol{\alpha}_{f}, \boldsymbol{z}_{f} \}$, defined over a window of $W$ frames, with $\boldsymbol{\theta}, \boldsymbol{\alpha}, \boldsymbol{z}$ representing the joint rotations, the root rotation, and the root translation, respectively.
Given this, we aim at meta-learning a personalized and optimal set of weights $\phi^*$ for the implicit human representation, effectively serving as a data-driven prior.
%
%
\par 
During the fine-tuning phase (Sec.~\ref{subsec:me_finetune}), we fine-tune the meta-learned weights $\phi^{*}$ to sparse in-the-wild images $\mathcal{{I}}_{f} = \{{I}_{f}^{k} | k\in [1;K_s] \}$ paired with masks ${\mathcal{O}}_{f} = \{{O}_{f}^{k} | k\in [1;K_s]\}$ with $K_s << K_d$. 
Notably, the motion $\mathcal{{M}}_{f,W}$ and surface dynamics of the performer as well as the lighting conditions and camera configuration during fine-tuning can significantly differ from the ones observed during training.
We later show that fine-tuning the meta-learned weights $\phi^{*}$ on $\mathcal{{I}}_f$ leads to faster and better convergence compared to random weights $\phi_0$.
%
%
\par
%
%
\vspace{-5pt}
\subsection{Meta-Learning a Reconstruction Prior}\label{subsec:me_bigpicture}
Meta-Learning~\cite{nichol2018first, finn2017model} trains a neural model in a way that makes it easier to adapt to novel tasks.
In our case, the novel tasks are sparse-view reconstruction and novel-view synthesis under unseen motions as well as lighting conditions.
For meta-learning the 3D geometry and appearance of the human, we represent the human as an implicit SDF and radiance field.
Given a ray $\mathbf{r}(\mathbf{o}, \mathbf{v})$ emanating from an image $\mathcal{I}_f$ with origin $\mathbf{o}$, and viewing direction $\mathbf{v}$, we sample $n$ points $\{ \mathbf{p}_{i} = \mathbf{o} + t_{i}\mathbf{v} | i\in[1;n] \}$ on depths $t_i$, and the implicit network $f_{\phi}$ predicts the color $\mathbf{c}_i$ and the SDF $s_i$ for each sample. 
We then utilize unbiased volume rendering~\cite{wang2021neus} along ray $\mathbf{r}$ to obtain RGB value $\hat{C}(\mathbf{r})$ and mask value $\hat{M}(\mathbf{r})$ :
\vspace{-5pt}
\small
\begin{gather}\label{eq:vol_rendering}
    \hat{C}(\mathbf{r}) = \sum_{i=1}^n T_i \alpha_i \mathbf{c}_i, \quad \hat{M}(\mathbf{r}) = \sum_{i=1}^n T_i \alpha_i, \quad  
    \alpha_i = \operatorname{max} \big( \frac{{\Psi}(s_{i}) - {\Psi}(s_{i+1})}{{\Psi}(s_{i})}, 0 \big),
\end{gather}
\normalsize
\vspace{0pt}
where $T_i$ is the accumulated transmittance, $\alpha_i$ is the unbiased weight function, and ${\Psi}(\cdot)$ represents the derivative of a sigmoid function.
These rendered rays can be used to supervise the implicit network $f_\phi(\cdot)$ using the following loss:
%
\vspace{-5pt}
\small
\begin{equation}
\begin{aligned}
\mathcal{L}(f_\phi,\mathbf{R}) = \lambda_\mathrm{c}\mathcal{L}_\mathrm{color} + \lambda_\mathrm{e}\mathcal{L}_{\mathrm{eik}} + \lambda_\mathrm{m}\mathcal{L}_{\mathrm{\mathrm{mask}}} + 
\lambda_\mathrm{s}\mathcal{L}_{\mathrm{sparse}},
\end{aligned} \label{eq:neus_loss}
\end{equation}
\normalsize
%
where $\mathbf{R}=\{\mathbf{r}_{k}| k\in[1;m]\}$ is a collection of $m$ rays, $\mathcal{L}_{\mathrm{color}}=\frac{1}{m}\sum_{k}\mathcal{H}({\hat{C}(\mathbf{r}_{k})},{C}(\mathbf{r}_{k}))$ is the Huber loss~\cite{huber1992robust}, 
$\mathcal{L}_{\mathrm{eik}}= %
\frac{1}{mn} \sum_{k, i} \left|1-\langle\mathbf{n}(\mathbf{p}_{k,i}),\left.\nabla_{\mathbf{p}_{k,i}} f_\phi(\mathbf{p}_{k,i})\right\rangle
\right|_{2}^{2}$ is the Eikonal loss~\cite{icml2020_2086} on the normal $\mathbf{n}(\mathbf{p}_{k,i})$ of the sample point $\mathbf{p}_{k,i}$.
$\mathcal{L}_\mathrm{mask}=\frac{1}{m}\sum_{k}BCE(\hat{M}{(\mathbf{r}_{k})}, {O}{(\mathbf{r}_{k})})$ is the binary cross entropy loss. $s(\mathbf{p}_{k, i})$ is the SDF value of point $\mathbf{p}_{k,i}$, and $\mathcal{L}_\mathrm{sparse}=\frac{1}{mn} \sum_{k, i} \exp^{-|s(\mathbf{p}_{k, i})|}$ is a sparseness regularization term~\cite{long2022sparseneus}.
${C}(\mathbf{r}_{k})$ and ${O}(\mathbf{r}_{k})$ are the ground truth color and mask values of ray $\mathbf{r}_{k}$.
To accelerate the rendering process, we parameterize the implicit field using a multi-resolution hashgrid~\cite{muller2022instant}.
Thus, the overall learnable parameters $\phi$ include the MLP weights of 
 the implicit function as well as the hashgrid parameters.
%
%
\par
Now, the initial weights $\phi_0$ for our coordinate-based implicit function $f_{\phi}$ can be learned through optimization-based meta-learning algorithms, like MAML~\cite{finn2017model} or Reptile~\cite{nichol2018first}.
Let us assume we have a dataset of tasks following the distribution $\mathcal{T}$, where a task $T \sim \mathcal{T}$ is defined as a set $\{\mathcal{L}, \{\mathbf{R}_{f}\sim \mathcal{I}_f, \mathcal{O}_f\}, f_\phi\}_{f}$ comprising the loss function, the input rays and the implicit function $f_{\phi}$. 
Our meta-learning optimization is performed in two nested loops. 
For the $j^\mathrm{th}$ iteration of the inner loop with sampled rays $\mathbf{R}^{j}$, it follows a gradient-descent optimization of the model parameters $\phi_j^{\prime}$ using a learning rate $l_{\mathrm{in}}$:
%
\small
\begin{equation}
\phi_j^{\prime} \leftarrow \phi_{j-1}^{\prime} - l_{\mathrm{in}}  \nabla_{\phi} \mathcal{L}(f_{\phi}, \mathbf{R}^{j})\vert_{\phi=\phi_{j-1}^{\prime}}.
\end{equation}
\normalsize
%
We leverage the model-agnostic meta-learning algorithm, Reptile~\cite{nichol2018first}, to optimize the initial weights by nesting an outer loop on top of the inner loop.
The outer loop then updates the meta-learned parameters $\phi_i$ through the first-order update equation:
%
\small
\begin{equation}\label{eq:reptile}
    \phi_{i} \leftarrow \phi_{i-1} + l_{\mathrm{out}}(\phi_M^{\prime}(\phi_{i-1}, T_i) - \phi_{i-1}),
\end{equation}
\normalsize
where $T_i$ are the tasks sampled at the $i^{th}$ outer iteration, $l_{\mathrm{out}}$ denotes the learning rate of outer loop. 
$\phi_M^{\prime}(\phi_{i-1}, T_i)$ represents the network weights initialized by the inner-loop with $\phi_{i-1}$, which are then optimized for $M$ steps on $T_i$.
The resulting weights, $\phi_N$, at the end of $N$ outer loop iterations serve as the optimal initial weights $\phi^{*}$ for the fine-tuning stage.
To keep the meta-learning computation tractable, we choose Reptile~\cite{nichol2018first} in Eq.~\ref{eq:reptile} as it uses a first-order approximation instead of MAML's second order gradient update.
%
%
\par 
Note that the above formulation directly meta-learns a volumetric rendering-based implicit network in the world-space.
This proves to be suboptimal (as shown in Fig.~\ref{fig:ablation} and Tab.~\ref{tab:ablation_geo}) as humans are highly articulated; a challenge further exacerbated by loose clothing.
In the task set $T$ consisting of humans in different body poses, the meta-learned hash encoding can map the same body point into different hash-table parameters.
To address this issue, we modify the volume rendering formulation in Eq.~\ref{eq:vol_rendering} by proposing our human template guidance.
%
%
\subsection{Template-guided Meta-Learning} \label{subsec:me_vr}
%
%

\textbf{Human Template.} 
Here, we introduce a generic motion-driven human template, defined as a deformable mesh $\bar{\mathbf{X}} \in \mathbb{R}^{V\times3}$, of a character in the canonical pose $\bar{\mathcal{M}}$ (as shown in Fig.~\ref{fig:overview}).
The template mesh $\bar{\mathbf{X}}$ can be deformed according to a given skeletal motion $\mathcal{M}_{f, W}$.
Each vertex in the human template first undergoes a per-vertex motion-dependent deformation in the canonical space through a transform $\mathbf{T}_{\mathrm{def}} (\mathcal{M}_{f, W}) \in \mathbb{R}^{4\times4}$.
After being deformed locally with $\mathbf{T}_{\mathrm{def}}$, one can perform forward kinematics $\mathbf{T}_{\mathrm{FK}}(\mathcal{M}_{f}, \bar{\mathcal{M}})$ to transform the deformed canonical space vertices $\mathbf{T}_{\mathrm{def}}\bar{\mathbf{X}}$ to world space vertices ${\mathbf{X}}$. 
The total invertible transformation from the template's canonical space to world space is: $\mathbf{T} = \mathbf{T}_{\mathrm{FK}}\mathbf{T}_{\mathrm{def}}$ and the template vertices are transformed as $\mathbf{X} = \mathbf{T} \bar{\mathbf{X}}$. 
%
%

\noindent\textbf{Template-guided Ray Warping.} 
The template mesh defined above provides us a coarse geometry information, which effectively guides our meta-learning since we are now able to learn the meta-weights in a pose-canonical space.
The sampled points $\mathbf{p}_{i}$ can be warped from the world space into the canonical space through our proposed template guided ray warping. 
Concretely, we first project each sampled point onto the nearest face (triangle) of the posed template mesh $\mathbf{X}$. 
To compute the transformation matrix $\mathbf{T}_{i}$, we barycentricaly interpolate the transformation of the vertices defining the nearest triangle.
Next, we apply the inverse transformation $\mathbf{T}_{i}^{-1}$ of the sampled point to obtain its canonical position $\bar{\mathbf{p}}_{i} = \mathbf{T}_{i}^{-1}\mathbf{p}_i$.
The canonical points, $\bar{\mathbf{p}_i}$, can now be encoded with the learnable hash encoding and rendered using the volumetric rendering approach described in Eq.~\ref{eq:vol_rendering}. 
Moreover the template allows us to perform empty skipping~\cite{hadwiger2017sparseleap} for point samples whose distance to the template exceeds a threshold $\eta$, thereby significantly accelerating the convergence.
In this way, our final meta-learning approach (see also Alg.~\ref{alg:reptile_neus}) is more efficient and effective compared to the baseline, i.e. learning in global space.
\vspace{-15pt}
\small
\begin{algorithm}
\caption{\model{}'s Meta-Learning Procedure}
\begin{algorithmic}
\small
  \Statex \textbf{Initialize}: Weights $\phi$, outer/inner learning rate $l^{out}$ and $l^{in}$
  \For{$i = 1, \ldots, N$}
    \State Sample frames $\{f^{(j)}\}_{j=1}^{M}$, views $\{k^{(j)}\}_{j=1}^{M}$ and generate rays $\{ {\mathbf{R}}^{(j)} \}_{j=1}^{M}$ 
    \State Set $\phi^{'}_0 = \phi$
    \For {$j = 1, \ldots, M$}
        \State Find closest triangle of points $\mathbf{p}$ and compute the point to face distance $d$
        \State Filter points with $d$  larger than threshold $\eta$
        \State Compute inverse transformation $\mathbf{T}$
        \State Transform points into canonical space $\bar{\mathbf{p}} = \mathbf{T}^{-1}\mathbf{p}$
        \State $\phi^{'}_{j} = \phi^{'}_{j-1} - l^{in} \nabla_{\phi}\mathcal{L}(f_{\phi}(\bar{\mathbf{R}}^{(j)})|_{\phi=\phi^{'}_{j-1}})$
    \EndFor
    \State $\phi \leftarrow \phi + l^{out} (\phi_M^{'} - \phi)$
  \EndFor
  \Statex \textbf{Result:} Optimal weights $\phi^* \leftarrow \phi$
\end{algorithmic}
\label{alg:reptile_neus}
\end{algorithm}
\normalsize
\vspace{-20pt}
%
%
\vspace{-5pt}
\subsection{Occlusion Handling and Fine-tuning} \label{subsec:me_finetune}
%
%
%
\begin{figure}[tb]
        \centering
	\includegraphics[width=0.9\linewidth]{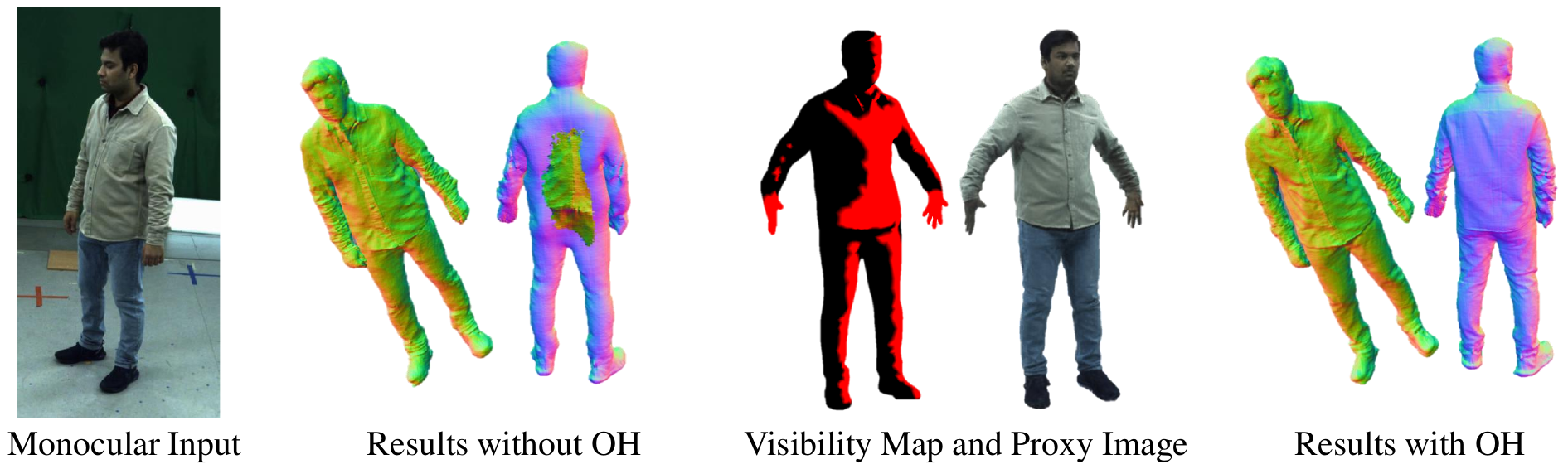}
     \vspace{-5pt}
	\caption
	{
        \textbf{Occlusion Handling.} 
        Our occlusion handling (abbreviated as OH) helps to regularize largely occluded regions.
	}
        \vspace{-10pt}
	\label{fig:occlusion_handling}
\end{figure}
%
%
%
At inference, when only the calibrated monocular or sparse-view RGB images $\mathcal{{I}}_f$, masks $\mathcal{{O}}_f$, and skeletal motion $\mathcal{{M}}_{f,W}$ are available, we initialize our implicit human field with the meta-learned weights $\phi^{*}$, conduct the space canonicalization with the fitted template $\mathbf{X}$ (Sec.\ref{subsec:me_vr}), and fine-tune using the loss from Eq.\ref{eq:neus_loss}. 
This can also be denoted as the task $\{\mathcal{L}, \{\mathbf{R}_{f} \sim {\mathcal{I}}_f, {\mathcal{O}}_f\}, f_{\phi}|{\phi_{0}}= \phi^*\}_{f}$.  
%
%

\noindent\textbf{Occlusion Handling.} 
For extreme occlusion cases, which may occur in the monocular setting (see also Fig.~\ref{fig:occlusion_handling}), we further propose a dedicated occlusion handling strategy by leveraging the posed human template.
Inspired by~\cite{sun2021neural}, we first pre-build an implicit human field on one frame of dense images $\mathcal{I}_{f}$ or multiple frames of (in-the-wild) sparse images ${\mathcal{I}}_f$, and then render virtual views of the canonical space as proxy RGB images $\mathcal{I}^\mathrm{proxy}$.
Next, during the fine-tuning stage, we determine per-vertex visibility by checking whether the vertex of the template is visible in any of the sparse input views.
Finally, in addition to sampling rays from the input images, we sample additional virtual rays for the occluded regions from $\mathcal{I}^\mathrm{proxy}$.
In practice, we found that the occlusion handling is only required in the case of monocular reconstruction (also see ablations).
%
%

%
%
\vspace{-10pt}
\section{Results} \label{sec:results}
\vspace{-5pt}
%
%
\textbf{Implementation Details.}
For meta-learning, we perform $M=24$ inner-loop gradient steps with $l^{out}=1.0$ and $l^{in}=1e-4$. 
The inner loop optimizer is Adam~\cite{kingma2017adam} while the outer one uses stochastic gradient descent.
Following NeuS~\cite{wang2021neus}, we perform a warmup in the inner loop at the beginning of training. 
If not stated otherwise, we use DDC~\cite{habermann2021} for space canonicalization, $\sim$80 cameras for meta learning, 4 cameras for fine-tuning in comparisons and ablations.
For more details, please refer to the supplemental material.
%
%

\noindent\textbf{Datasets.} 
%
%
%
\begin{figure}[tb]
        \centering
	\includegraphics[width=0.99\linewidth]{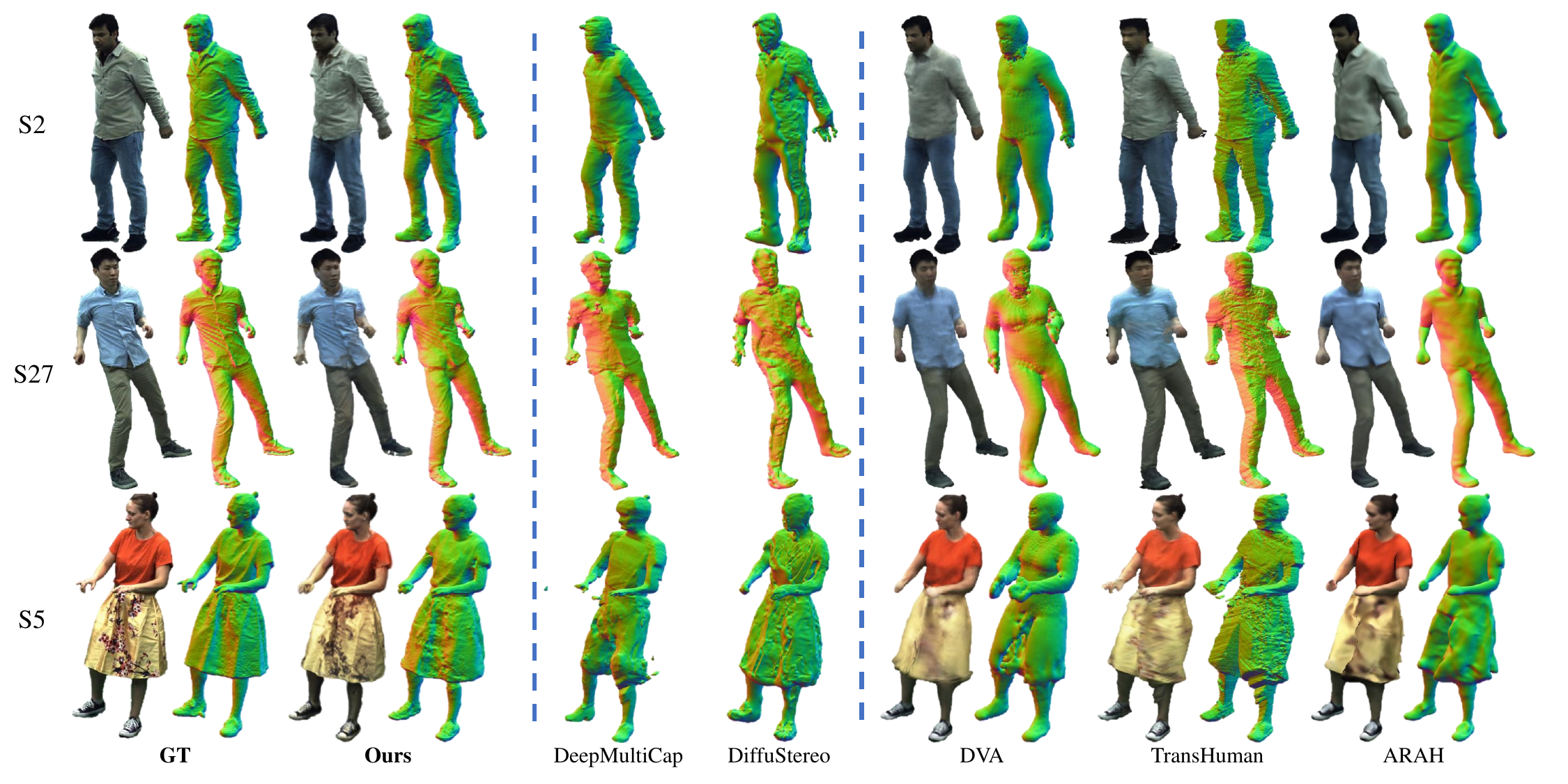}
        \vspace{-10pt}
	\caption
	{
        \textbf{Qualitative Comparison.}
        Compared to previous works, our method better captures high-frequency geometry on cloth wrinkles and faithfully recovers appearance details such as facial details and clothing textures.
	}
    \vspace{-10pt}
	\label{fig:comparison}
\end{figure}
%
%
\begin{table}[t]
   \centering
     \caption{
  \textbf{Quantitative Comparison.}
  Our method achieves state-of-the-art results for novel-view synthesis and geometry reconstruction.
  *Note, that ARAH requires 4D scans for meta learning and videos for the fine-tuning whereas other methods solely require static images.}
  \vspace{-10pt}
    \scalebox{0.73}{
    \begin{tabular}{|c|c|ccc|ccccc|}
    \hline 
    \multirow{2}{*} { \textbf{Method} } &  \multirow{2}{*} { \textbf{Subject} }    & \multicolumn{3}{c|}{\textbf{Appearance} } & \multicolumn{5}{c|}{ \textbf{Geometry} } \\
    \cline{3-10} & & \textbf{PSNR} $\uparrow$ & \textbf{SSIM} $\uparrow$ & \textbf{LPIPS} $\downarrow$  & \textbf{NC-Cos} $\downarrow$ & \textbf{NC-L2} $\downarrow$ & \textbf{Chamfer} $\downarrow$ & \textbf{P2S} $\downarrow$ & \textbf{IOU} $\uparrow$  \\
    \hline 
    \multirow{3}{*} { DeepMultiCap ~\cite{zheng2021deepmulticap} }
    & S2 & - & - & - & 0.143 & 0.445 & 1.305 & 1.303 & 0.764  \\
    & S27 & - & - & - & 0.149 & 0.454  & 1.466  & 1.831 & 0.797  \\
    & S5 & - & - & - & 0.183 & 0.504  & 2.238 & 2.224 & 0.794  \\
    \hline 
    \multirow{3}{*} { DiffStereo ~\cite{shao2022diffustereo}}
     & S2 & - & - & -  & 0.150 & 0.450 & \bronze{1.150} & \bronze{1.261} & \bronze{0.813} \\
    & S27 & - & - & -  & 0.161 & 0.471 & \bronze{1.093} & \bronze{1.185} & \bronze{0.818}  \\
    & S5 & - & - & -  & 0.173 & 0.491 & \bronze{1.714} & \silve{1.787} & \bronze{0.805}  \\
    \hline 
    \multirow{3}{*} { DVA ~\cite{remelli2022drivable} }
    & S2 & 25.469 & \bronze{0.817} & 0.307 & \bronze{0.123} & \bronze{0.402} & 1.647 & 2.245 & 0.426 \\
    & S27 & \silve{24.709} & \bronze{0.826} & \bronze{0.295} & \bronze{0.137} & \bronze{0.429} & 1.624 & 2.145 & 0.399  \\
    & S5 & \bronze{23.093} & \bronze{0.750} & \bronze{0.344} & \bronze{0.167} & \bronze{0.482} & 2.308 & 2.805 & 0.247  \\
    \hline 
    \multirow{3}{*} { TransHuman ~\cite{Pan_2023_ICCV} } 
    & S2 & \bronze{25.770} & 0.810 & \bronze{0.305} & 0.150 & 0.445 & 1.618 & 2.238 & 0.767 \\
    & S27 & 23.876 & 0.800 & 0.304 & 0.146 & 0.440 & 1.487 & 1.970 & 0.791  \\
    & S5 & 23.072 & 0.736 & 0.349 & 0.183 & 0.495 & 1.898 & 2.039 & 0.763 \\
    \hline 
    \multirow{3}{*} { ARAH* ~\cite{ARAH:ECCV:2022}}
    & S2 & \silve{26.279} & \silve{0.833} & \silve{0.302} & \gold{0.079} & \gold{0.315} & \silve{0.839} & \silve{0.913} & \silve{0.859}  \\
    & S27 & \bronze{24.666} & \silve{0.841} & \silve{0.282} & \gold{0.080} & \gold{0.316} &\silve{0.776}& \silve{0.815} & \silve{0.850}  \\
    & S5 & \silve{23.532} & \silve{0.775} & \silve{0.332} & \gold{0.139} & \gold{0.419} & \silve{1.620} & \bronze{1.915} & \silve{0.842} \\
    \hline 
    \multirow{3}{*} { \textbf{Ours} }
    & S2 & \gold{26.529} & \gold{0.841} & \gold{0.249} & \silve{0.096} & \silve{0.351} & \gold{0.679} & \gold{0.814} & \gold{0.887} \\
    & S27 & \gold{25.284} & \gold{0.849} & \gold{0.247} & \silve{0.104} & \silve{0.370} & \gold{0.614} & \gold{0.734} & \gold{0.891} \\
    & S5 & \gold{23.996} & \gold{0.777} & \gold{0.302} & \silve{0.147} & \silve{0.448} & \gold{1.133} & \gold{1.277} & \gold{0.877} \\
    \hline
    \end{tabular}
    }
  \vspace{-10pt}
  \label{tab:quantitative}	
  %
\end{table}
We evaluate our approach on the publicly available dense-view dome dataset DynaCap~\cite{habermann2021} and our \dataset{} dataset. 
DynaCap provides around 100-view videos, foreground masks, and motions.
We choose two subjects wearing loose (S5) and tight clothing (S3 in the supplemental). 
To compensate for the lack of in-the-wild sequences in DynaCap, we collect a new dataset \dataset{}. 
It contains 110 cameras to capture the performer inside a dome, and a movable 5-camera setup to record videos in different in-the-wild scenarios. 
We captured two subjects in two camera setups. 
We recover the skeletal motion using markerless motion capture~\cite{stoll2011fast} using 34 views for the dome setups and 5 views for the in-the-wild sequences.
For all experiments and methods, we use this motion (if not stated otherwise) as this work focuses on surface recovery rather than motion capture.
We also provide ablation on motions only from sparse  cameras in the supplemental. 
%
%

\noindent\textbf{Metrics and Ground-Truth.} 
To obtain the ground truth geometry for the dome captures, we leverage the recent state-of-the-art implicit reconstruction method InstantNSR~\cite{zhao2022human} trained on the \textit{dense} RGB views.
To evaluate the rendering quality, we report the peak signal-to-noise ratio (PSNR), structural similarity index (SSIM), and learned perceptual image patch similarity (LPIPS)~\cite{zhang2018unreasonable}. 
To evaluate geometry results, we report Chamfer distance (CD), point-to-mesh distance (P2S), intersection over union (IOU), cosine normal consistency (NC-Cos), and L2 normal consistency (NC-L2).
For all metrics, we sample every 100\textit{th} frame from the test set, where the actor is performing unseen motions, and report the average. We provide an illustration of input cameras and evaluation cameras distribution in the supplementary material. 
%
%
\vspace{-10pt}
\subsection{Comparisons} \label{sec:comparison}
\vspace{-10pt}
\begin{table}[t]
  \centering
        \caption{
      \textbf{Quantitative Ablation.}
      Here, we evaluate different initialization strategies and canonicalization types on rendering quality. 
      Networks initialized  with random or pre-trained weights tend to overfit the inputs, while our meta weights generalize well to novel views.
      }
        \vspace{-10pt}
      \scalebox{0.65}{
    \begin{tabular}{|c|c|c|c|c|c|c|c|c|c|c|c|}
    \hline   \multirow{2}{*} { \textbf{Init Type} } &  \multirow{2}{*} { \textbf{Cano Type} } & \multicolumn{5}{c|}{ \textbf{Input View} } & \multicolumn{5}{c|}{\textbf{Novel View}} \\
    \cline{3-12} %
    &  & \textbf{PSNR} $\uparrow$ & \textbf{SSIM} $\uparrow$  & \textbf{LPIPS} $\downarrow$ & \textbf{NC-Cos} $\downarrow$ &  \textbf{NC-L2} $\downarrow$ & \textbf{PSNR} $\uparrow$  & \textbf{SSIM} $\uparrow$  & \textbf{LPIPS} $\downarrow$ & \textbf{NC-Cos} $\downarrow$ &  \textbf{NC-L2} $\downarrow$ \\
    \hline 
    \multirow{3}{*} { Random }
     & Root  & \silve{31.678} & \silve{0.955} & \gold{0.113} & 0.173 & 0.455 & 23.802   & 0.782   & 0.276  & 0.171   & 0.454 \\
     \cline{2-12} 
    & SMPL & 31.180 & 0.953 & \bronze{0.120} & 0.225 & 0.535 & 23.691    & 0.770   & 0.292 &	0.214 	& 0.523\\
    \cline{2-12} 
    & DDC  & \bronze{31.429} & \bronze{0.954}  & \bronze{0.120} &	0.223 & 0.530 & 23.566   & 0.772   & 0.289 &0.216 	 & 0.523\\
    \hline 
    \multirow{3}{*} { Pretrain }
    & Root  & - & - & - & - & - & - & - & - & - & - \\
    \cline{2-12}  
    & SMPL  & - & - & - & - & - & - & - & - & - & -\\
    \cline{2-12}  
    & DDC  & \gold{32.423} & \gold{0.959} & \silve{0.114} & \bronze{0.110}	& \bronze{0.386} & \bronze{25.829}   & \bronze{0.820}   & \gold{0.239} & \silve{0.116}		 &	\bronze{0.395} \\
    
    \hline 
    \multirow{3}{*} {\textbf {Meta} }
    & Root  & 24.477 & 0.870 & 0.253 & 0.170	&0.480  & 23.094   & 0.764  & 0.326 & 0.168			& 0.475 \\  \cline{2-3}
    \cline{2-12} 
    & SMPL  & 28.224 & 0.903 & 0.215 &	\silve{0.092} &	\silve{0.345} & \silve{26.120}   & \silve{0.833}  & \bronze{0.258} & \gold{0.096}   	& \gold{0.349} \\
    \cline{2-12} 
    & \textbf{ DDC }  & 29.176 & 0.912 & 0.209 & \gold{0.089}	&	\gold{0.341} & \gold{26.529}   & \gold{0.841}   & \silve{0.249} & \gold{0.096} 	& \silve{0.351} \\
    \hline  
    \end{tabular}
    }
    \vspace{-20pt}
  \label{tab:ablation_appearance}
  %
\end{table}
%
%
\begin{figure}[t]
        \centering
	\includegraphics[width=0.95\linewidth]{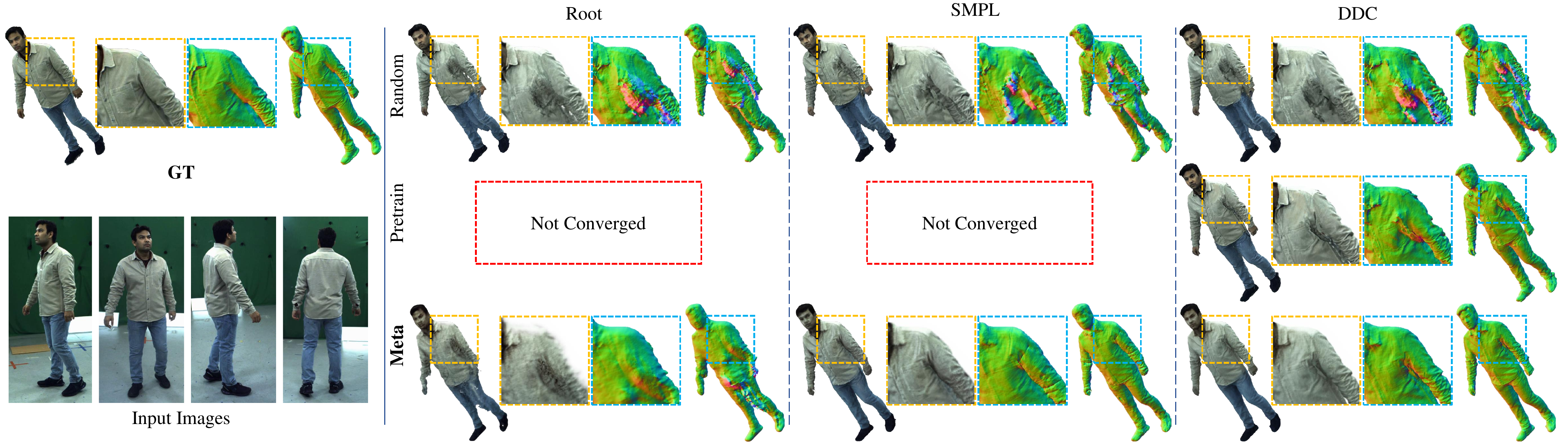}
        \vspace{-10pt}
	\caption
	{
        \textbf{Qualitative Ablation.}
        We show results for different weight initialization strategies and space canonicalization types. 
        Our meta-learning paradigm, together with the proposed space canonicalization, achieves the best result.
        We highlight that this applies irrespective of the choice of the character model (SMPL~\cite{SMPL:2015} or DDC~\cite{habermann2021}).
	}
        \vspace{-15pt}
	\label{fig:ablation}
\end{figure}

%
%

%
We compare our method with recent reconstruction and novel synthesis methods providing each of them wide-baseline four camera views as input during inference.
Specially, we compare our method with generalizable reconstruction~\cite{zheng2021deepmulticap, shao2022diffustereo} and rendering~\cite{Pan_2023_ICCV} methods, a subject-specific rendering method~\cite{remelli2022drivable}, and a subject-specific optimization method with geometry prior initialization~\cite{ARAH:ECCV:2022}. 
For more details, please refer to supplementary materials.
\par 
\noindent\textbf{(1) Generalizable Methods.} 
Given large-scale 3D human datasets~\cite{zheng2021deepmulticap, twindom}, DeepMultiCap~\cite{zheng2021deepmulticap} learns to predict occupancy fields from multi-view RGBs by fusing RGB features, normal features and voxel features with an attention layer, while DiffuStereo~\cite{shao2022diffustereo} trained a diffusion model to refine coarse stereo disparity.
TransHuman~\cite{Pan_2023_ICCV} captures global relationships of human parts in the canonical space and learns to predict radiance fields from multi-view videos.
%
%
\par 
\noindent\textbf{(2) Subject-specific Methods.}
Given dense-view RGB videos, DVA~\cite{remelli2022drivable} learns articulated volumetric primitives attached to a parametric body model combined with texel-aligned features. 
ARAH~\cite{ARAH:ECCV:2022} extends a meta-learned SDF prior~\cite{MetaAvatar:NeurIPS:2021} with a root-finding method to reconstruct clothed avatars from a sparse set of multi-view RGB videos.
%
%
\par
We demonstrate qualitative and quantitative comparisons in Fig.~\ref{fig:comparison} and Tab.~\ref{tab:quantitative}. 
\begin{wraptable}{l}{0.5\linewidth}
  \centering
  \vspace{-10pt}
    \caption{
          \textbf{Quantitative Ablation.}
          Here, we study initialization strategies and canonicalization types in terms of geometry results.
          Again, our meta-learning paradigm in combination with space canonicalization outperforms the baselines.
          }
    \vspace{10pt}
      \small
          \scalebox{0.8}{
    \begin{tabular}{|c|c|c|c|c|}
    \hline
    \textbf{Init Type}  &   \textbf{Cano Type}  & \textbf{Chamfer}$\downarrow$ & \textbf{P2S}$\downarrow$ & \textbf{IOU}$\uparrow$ \\
    \hline 
    \multirow{3}{*} { Random }
     & Root 
     & 1.610 & 2.253 & 0.823 \\
     \cline{2-5} 
    & SMPL 
    & 1.291 &	1.596 &	0.806 \\
    \cline{2-5}
    & DDC 
    & 1.184	& 1.498 &	0.815  \\
    \hline 
    \multirow{3}{*} { Pretrain }
    & Root
    & - & - & -   \\
    \cline{2-5}  
    & SMPL 
    &- & - & -   \\
    \cline{2-5}  
    & DDC 
    & \silve{0.726}	& \silve{0.855}	& \bronze{0.882}  \\
    
    \hline 
    \multirow{3}{*} { \textbf{Meta }} 
    & Root 
    & 1.692	& 2.001	& 0.817	 \\ 
    \cline{2-5}  
    & SMPL 
    & \bronze{ 0.799}	& \bronze{0.897	}& \silve{0.885}  \\
    \cline{2-5}  
    & \textbf{ DDC }
    & \gold{ 0.679 } &	\gold{0.814} & \gold{0.887} \\
    \hline
    \end{tabular}
    }
  \vspace{-20pt}
  \label{tab:ablation_geo}
  \normalsize
\end{wraptable}
DeepMultiCap~\cite{zheng2021deepmulticap} directly regresses occupancy in world space, due to the articulated structure of humans and the limited scale of 3D scan datasets, their generalization ability is rather limited.
DiffuStereo~\cite{shao2022diffustereo} struggles to find correct image correspondences under our challenging wide-baseline camera setup, leading to noisy reconstruction.
DVA~\cite{remelli2022drivable} learns rotation, translation, and scale of primitives to recover rough geometry. 
However, their texture un-projection step is sensitive to the template, and tracking errors of the template lead to blurred results.
TransHuman~\cite{Pan_2023_ICCV} integrates pixel-aligned features to improve texture details. 
However, they do not model geometry explicitly, which result in noisy geometry and texture boundaries.
ARAH's geometry network~\cite{MetaAvatar:NeurIPS:2021} learns a meta prior taking 3D position as input and outputs the SDF value.
%
%
%
\begin{figure}[tb]
    \centering
    \includegraphics[width=0.68\linewidth]{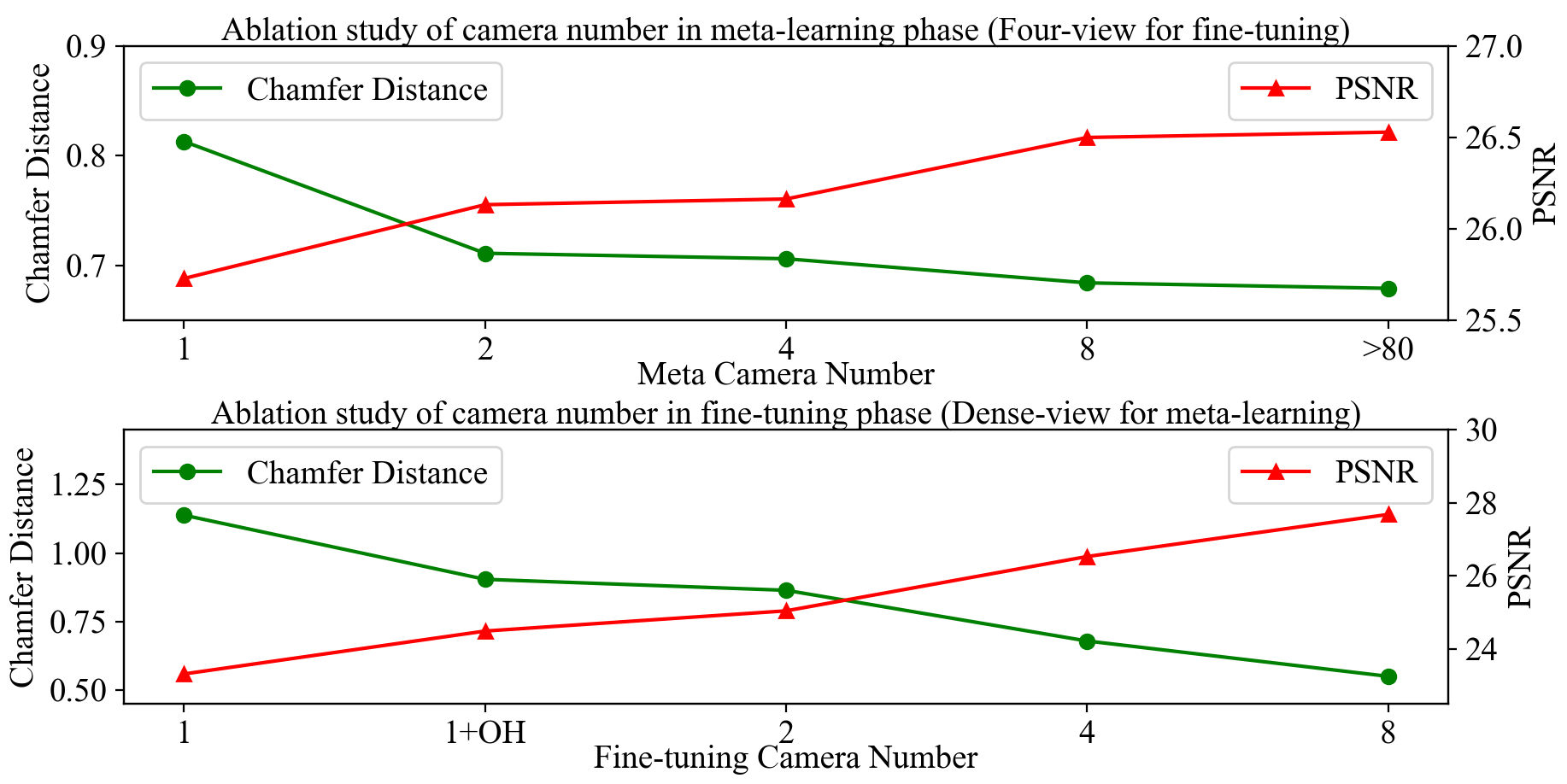}
 \vspace{-5pt}
	\caption
	{
    \textbf{Qualitative Ablation.}
    Here, we study the influence of occlusion handling and view numbers during meta-learning (top) and fine-tuning (bottom).
	}
	\label{fig:camera_number}
	%
  \vspace{-10pt}
\end{figure}
%
%
However, due to their purely MLP-based architecture and non-end-to-end design, we found their results suffer from blurred appearance and less detailed geometry.
Besides, the root-finding is not efficient taking over ten hours with 4 GPUs to finetune.
In contrast, we use explicit hash encodings for spatial features, which is efficient and effective for meta feature learning. 
All the rendering methods above use parametric models~\cite{SMPL:2015, SMPL-X:2019} as template, so they can not model loose clothing correctly (see S5 in Fig.~\ref{fig:comparison}).
In contrast, our method recovers high-fidelity appearance and geometry even for loose types of apparel in minutes under this challenging setup, i.e. 4 wide-baseline cameras. 
%
%
\subsection{Ablation Studies} \label{sec:ablation}
We perform ablation studies on the test set of the ``S2'' subject to evaluate the effectiveness of our design choices.
%
%
%

\noindent\textbf{Weight Initialization and Space Canonicalization.} 
%
%
\begin{figure}[tb]
     \centering
	\includegraphics[width=0.8\linewidth]{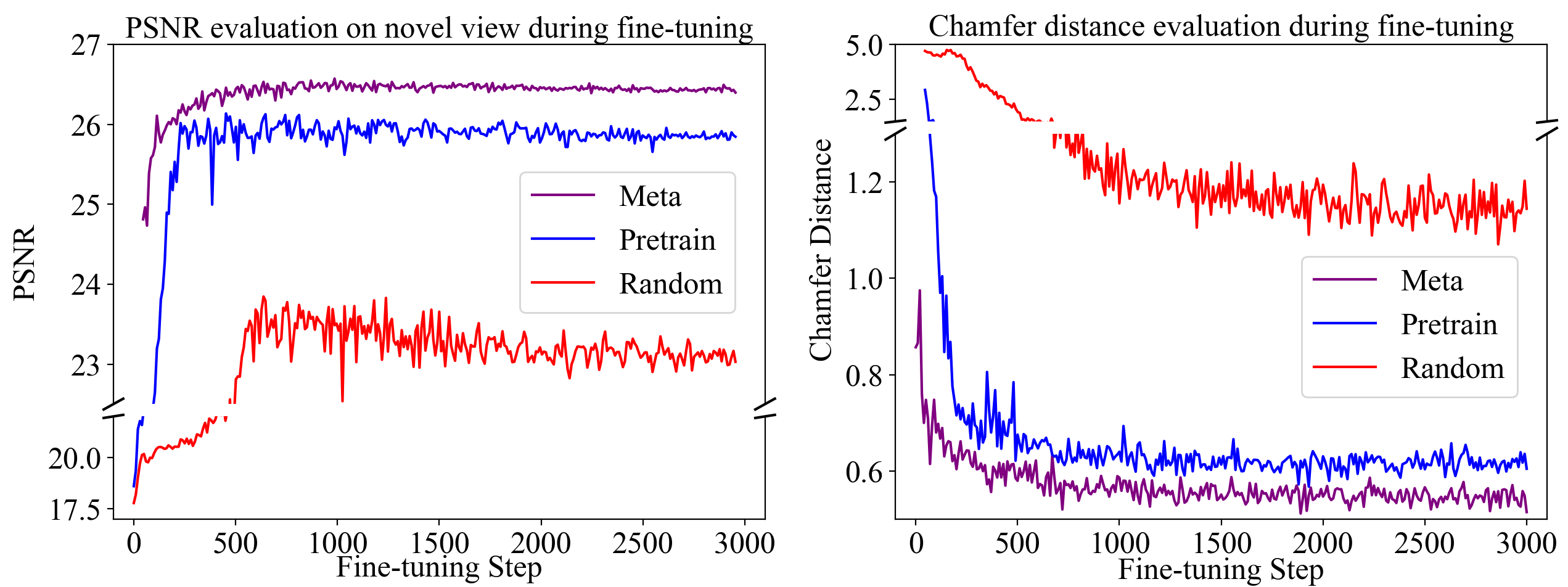}
	%
        \vspace{-5pt}
        \caption
        {
        \textbf{Convergence and Quality.}
        Note that our meta-learned weights achieve faster convergence and higher accuracy.
	}
	\label{fig:convergence}
        \vspace{-20pt}
\end{figure}
%
%
Tab.~\ref{tab:ablation_appearance}/\ref{tab:ablation_geo} and Fig.~\ref{fig:ablation} show the ablation results for weight initialization and space canonicalization.
We choose the following baselines for weight initialization: 
\textit{Random}: Network weights are initialized randomly. 
\textit{Pretrain}: Network weights are pre-trained on all training frames. 
\textit{Meta}: Network weights are learned through meta-learning. 
For space canonicalization, we choose the following baselines:
\textit{Root}: The space transformation is defined by the skeletal root rotation and translation.
\textit{SMPL}: The space transformation is defined by the nearest surface point on the SMPL~\cite{SMPL:2015} template.
\textit{DDC}: The space transformation is defined by the nearest point on the deformable DDC~\cite{habermann2021} template.\\
%
%
%
%
Random initialization typically overfits the input views, however, on novel views it performs poorly.
Meta-learning in root-normalized space also performs poorly as the articulated nature of the humans poses a significant challenge for meta-learning the optimal network weights.
In contrast, the combination of our key technical components, i.e. meta-learning weights \textit{and} space canonicalization, effectively addresses those issues and achieves the best result for, both, geometry reconstruction and novel view synthesis.
Moreover, we highlight that this holds true irrespective of the specific choice of the human template, i.e. SMPL or DDC.
%
%

\noindent\textbf{Number of Camera Views and Occlusion Handling.}
Since our method solely takes videos for meta-learning and images for fine-tuning, it naturally supports arbitrary camera setups in \textit{both} stages.
As illustrated in Fig.~\ref{fig:camera_number}, when increasing the number of cameras, our method obtains better rendering and reconstruction results. 
While this is not too surprising, we highlight that even under sparser settings our method performs reasonably well.
Most interestingly, meta-learning can even be performed on a monocular video, which is not possible for ARAH~\cite{ARAH:ECCV:2022} and MetaAvatar~\cite{MetaAvatar:NeurIPS:2021}, since they rely on scans for meta-learning.
Besides, at inference, our method allows fine-tuning using just a single view.
Moreover, when performing monocular fine-tuning, our occlusion handling approach (1+OH) outperforms the baseline, which is not using this component.
%
%
%

%
%
\begin{figure}[tb]
        \centering
	\includegraphics[width=0.9\linewidth]{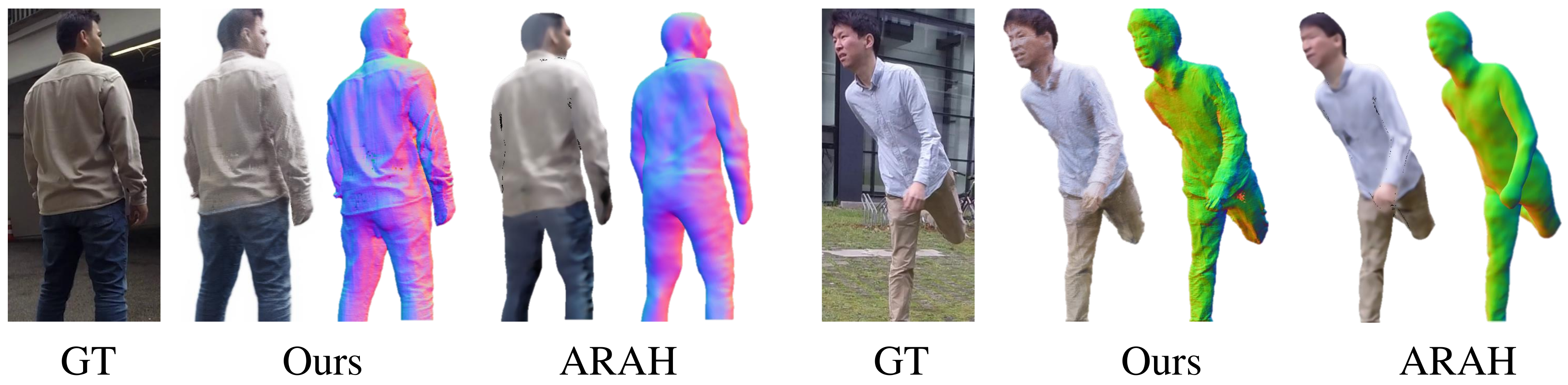}
 \vspace{-5pt}
	\caption
	{
        \textbf{Qualitative In-the-wild Comparison.} 
        To evaluate the robustness to in-the-wild conditions, we compare our work and ARAH on \dataset{}.
        Our approach achieves significantly better results in terms of appearance and geometric details.
	}
	\label{fig:ITW}
          \vspace{-15pt}
\end{figure}
%
%
\noindent\textbf{Convergence Speed and Quality.} 
In Fig.~\ref{fig:convergence}, we evaluate the convergence of our method compared to random and pre-trained weight initializations.
Our proposed design outperforms the baselines in, both, convergence speed and accuracy.
Since one fine-tuning step only takes $50ms$, our meta-learned initialization already converges after 40$s$.
%
%
%
\subsection{Evaluation on In-the-wild Sequences} \label{sec:in_the_wild}
%
\begin{wraptable}{l}{0.45\linewidth}
   \centering
    \vspace{-30pt}
       \caption{
    \textbf{Quantitative In-the-Wild Comparison.} 
    We outperform ARAH on in-the-wild sequences.
    } 
   \small
    \scalebox{0.7}{
    \begin{tabular}{|c|c|c|c|c|}
    \hline 
    \multirow{1}{*} { \textbf{Method} } &  \multirow{1}{*} { \textbf{Subject} } & \textbf{PSNR} $\uparrow$ & \textbf{SSIM} $\uparrow$ & \textbf{LPIPS} $\downarrow$    \\
    \hline 
    \multirow{2}{*} { ARAH* ~\cite{ARAH:ECCV:2022}}
    & S2 & \silve{19.027} & \silve{0.608} & \silve{0.412}   \\
    & S27 & \silve{22.075} & \silve{0.702} & \silve{0.343}   \\
    \hline 
    \multirow{2}{*} { \textbf{Ours} }
    & S2 & \gold{19.156} & \gold{0.661} & \gold{0.351}  \\
    & S27 & \gold{22.080} & \gold{0.750} & \gold{0.260}  \\
    \hline
    \end{tabular}
    }
    %
    \vspace{-30pt}
    \label{tab:ITW}	
      \normalsize
\end{wraptable}
Fig.~\ref{fig:ITW} and Tab.~\ref{tab:ITW} show results on our \dataset{} dataset.
Our method recovers high-fidelity geometry and appearance, consistently outperforming ARAH. 
These results show the robustness of our method to complex background, lighting conditions, and camera types.
%

%
%
\section{Conclusion} \label{sec:conclusion}
%
This paper introduced \model{}, a novel approach towards high-fidelity performance capture and photo-realistic rendering of humans from very sparse multi-view or even monocular RGB images.
We introduced a hybrid representation that benefits from both explicit and implicit human representation.
Through comprehensive evaluations, we demonstrated that performing meta-learning with space canonicalization proves to be a crucial design factor, thereby providing a strong data-driven prior on the human capture.
Our results demonstrate high-fidelity human reconstruction and free-view rendering as well as the versatility of approach in terms of camera setups, clothing types, and canonicalization strategies.
In the future, we plan to explore even faster, potentially real-time, fine-tuning strategies as well as learning priors across individuals.
\par
\noindent\textbf{Limitations.} 
Although our method achieves high-fidelity personalized human reconstruction and rendering from sparse observation, it still has a few limitations. First, our method can be sensitive to template fitting or motion capture results. Second, we do not take the temporal information into consideration. Integrating reliable observations or constraints from adjacent frames like Newcombe et al.~\cite{newcombe2015dynamicfusion} may lead to more robust geometry and rendering results. Third, our method can not handle hands very well. Modeling hand with a  more fine-grained template and motion capture could be a potential solution. 
%

\section*{Acknowledgements}
This research was supported by the ERC Consolidator Grant 4DRepLy (770784).


%
%
\bibliographystyle{splncs04}
\bibliography{main}
\clearpage

%
%
\section{Overview of Supplementary Material}
\vspace{-3pt}
To facilitate a more comprehensive analysis and understanding of \model{} and experiment configurations, we offer additional results (Sec.~\ref{sec:more_results}), method details (Sec.~\ref{sec:template_model},~\ref{sec:_supp_image_proxy}), implementation details (Sec.~\ref{sec:supp_implementation}), method costs (Sec.~\ref{sec:supp_cost}), comparisons (Sec.~\ref{sec:supp_comparison_s3},~\ref{sec:supp_comparison_s3_mono}), ablations (Sec.~\ref{sec:supp_ablation_mocap}), applications (Sec.~\ref{sec:supp_applications}), and temporal results (Sec.~\ref{sec:supp_temporal}).
%
%
\begin{figure}
	\includegraphics[width=\linewidth]{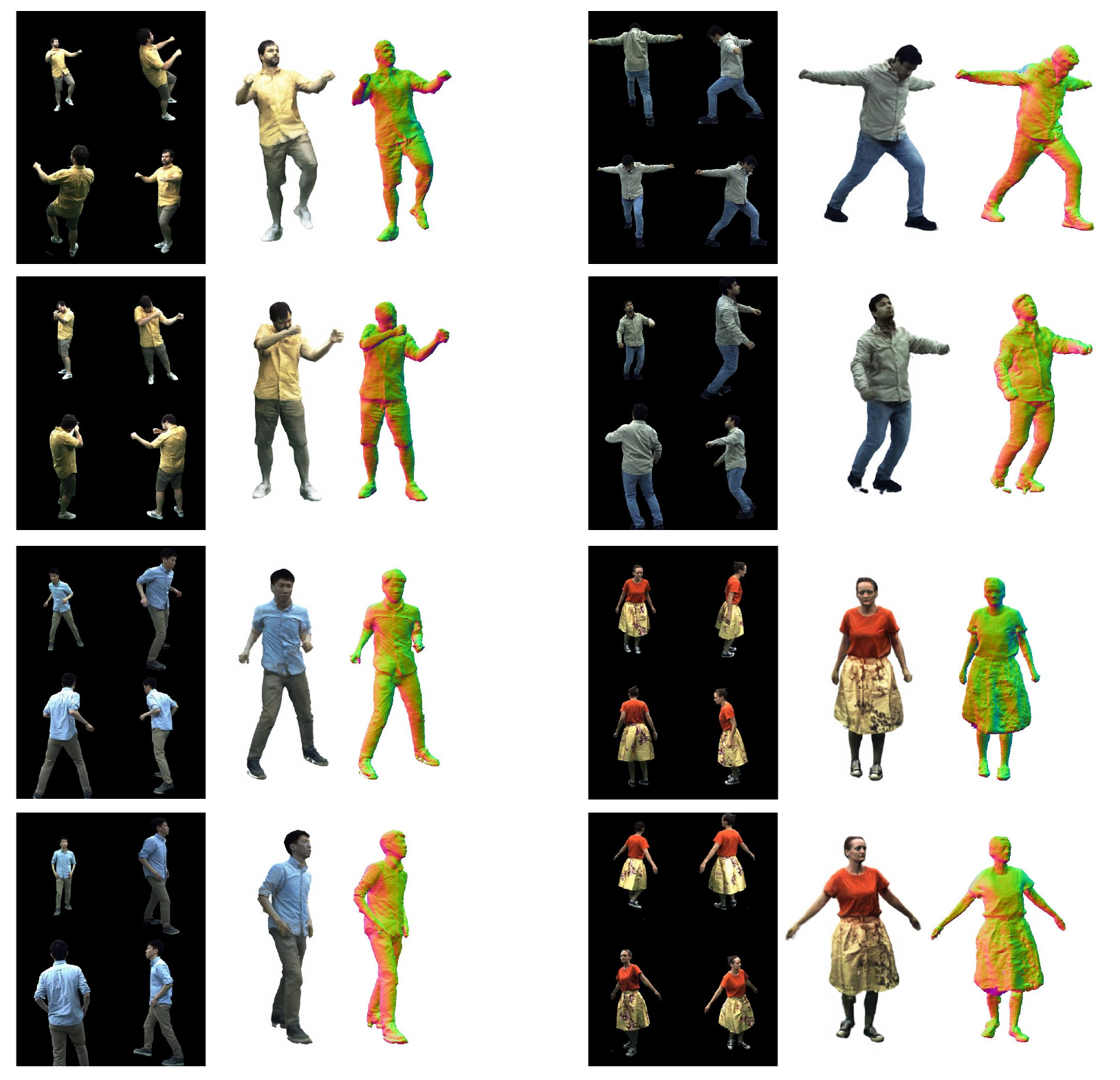}
	\caption
	{
	    \textbf{Qualitative Results.} Here, we showcase additional qualitative results of \model{} utilizing four-view images as inputs, demonstrating its robustness across diverse poses and different subjects.
	}
	\label{fig:more_results}
        \vspace{-20pt}
\end{figure}
%
%

\section{More Results on Different Poses and Subjects} \label{sec:more_results}
Fig.~\ref{fig:more_results} presents additional qualitative results showcasing the performance of our method across diverse motions and subjects. Since our method learns the meta prior in the canonical pose space, it is robust to various testing poses.

\section{Template Model} \label{sec:template_model} 
We revisit two types of human template, SMPL\cite{SMPL:2015} and DDC\cite{habermann2021} and demonstrate how to compute the transformation matrix and deformed position for each vertex, which are crucial for the canonicaliztion step. Here, each template has vertices $\bar{\mathbf{X}} \in \mathbb{R}^{V\times3}$ in canonical pose $\bar{\mathcal{M}}=\{ \bar{\boldsymbol{\theta}}, \bar{\boldsymbol{\alpha}}, \bar{\boldsymbol{z}} \}$. $\bar{\boldsymbol{\theta}}, \bar{\boldsymbol{\alpha}}, \bar{\boldsymbol{z}}$ represents the joint rotations, the root rotation, and the root translation of canonical pose, respectively. We define a window size $W$ skeletal motion at time $f$ as $\mathcal{M}_{f,W} = \{\boldsymbol{\theta}_{f-W}, \boldsymbol{\alpha}_{f-W}, \boldsymbol{z}_{f-W}, ..., \boldsymbol{\theta}_{f}, \boldsymbol{\alpha}_{f}, \boldsymbol{z}_{f} \}$. If $W$ is not explicitly set, it defaults to 1.

\subsection{Parametric Template--SMPL} SMPL~\cite{SMPL:2015} is a parametric human body model. 
It characterizes $V$ vertices and $J$ joint positions of the human mesh using shape parameters, $\mathbf{\beta}$, canonical pose parameters, $\bar{\mathcal{M}}$, and pose parameters, $\mathcal{M}$. 
The overall mesh deformation, $\mathbf{T}_{\mathrm{def}}(\mathbf{\beta},\mathcal{M})$, is determined by the sum of shape dependent displacements and pose dependent displacements.
Linear Blend Skinning (LBS) is employed for animating the deformed mesh:
\begin{align}
    \mathbf{T}_{\mathrm{FK},i}(\mathcal{M}, \bar{\mathcal{M}}) &= \sum_{j=1}^{J}w_{j,i}T_{j}(\mathcal{M}) (\sum_{j=1}^{J}w_{j,i}T_{j}(\bar{\mathcal{M}}))^{-1} \\
    \mathbf{T} &= \mathbf{T}_{\mathrm{FK}}(\mathcal{M}, \bar{\mathcal{M}})\mathbf{T}_{\mathrm{def}}(\mathbf{\beta},\mathbf{\mathcal{M}})\\
    \mathbf{X} &= \mathbf{T} \bar{\mathbf{X}}
\end{align}
where $w_{j,i}$ is the blend weight from joint $j$ to vertex $i$, $T_{j}(\mathcal{M})$ denotes the joint $j$'s local transformation, $\mathbf{T}_{\mathrm{FK},i}(\mathcal{M}, \bar{\mathcal{M}})$ represents the global transformation of the deformed vertex $i$ from canonical pose $\bar{\mathcal{M}}$ to the pose $\mathcal{M}$.

\subsection{Deformable Template--DDC}
DDC\cite{habermann2021} is a personalized deformable body model. 
It models motion-dependent body deformation with embedded graph deformation and vertex displacements. Given a window size $W$ skeletal motion $\mathcal{M}_{f,W}$ at time $f$, it employs Graph Convolutional Networks (GCN)~\cite{zhang2019graph} to estimate the embedded-graph's deformation parameters $\mathbf{A}, \mathbf{T} \in \mathbb{R}^{K \times 3}$ and the per-vertex displacements $d$. The final geometry is obtained through Dual Quaternion Skinning (DQS):

\begin{equation}
\mathbf{T}_{\mathrm{def},i}(\mathcal{M}_{f,W}) = \sum_{k} \mathrm{w}_{k,i} \left[\begin{array}{c|c}
    R(\mathbf{a}_k) & d_{i} + (I - R(\mathbf{a}_k))\mathbf{g}_k + \mathbf{t}_k \\
    \hline 
    \overrightarrow{0} & 1
\end{array}\right]
\end{equation}

\begin{align}
    \mathbf{T} &= \mathbf{T}_{\mathrm{FK}}(\mathcal{M}_{f}, \bar{\mathcal{M}})\mathbf{T}_{\mathrm{def}}(\mathcal{M}_{f,W})\\
    \mathbf{X} &= \mathbf{T} \bar{\mathbf{X}}
\end{align}
where $\mathrm{w}_{k,i}$ represents the weight from graph node $k$ to vertex $i$, $R(\cdot)$ transforms Euler representations into matrix representations, $\mathbf{a}_k$ and $\mathbf{t}_k$ are node $k$'s local rotation and translation, $\mathbf{g}_k$ denotes the position of node $k$. $\mathbf{T}_{\mathrm{FK}}(\mathcal{M}_{f}, \bar{\mathcal{M}})$ represents global transformation from initial pose $\bar{\mathcal{M}}$ to pose $\mathcal{M}_{f}$. In the implementation of DDC, the window size is set to 3. Note that, $\mathcal{M}_{f}$ represents the pose at frame $f$, and the motion window of it is 1.

\section{Proxy Image Generation for Occlusion Handling} \label{sec:_supp_image_proxy}
%
%
\begin{figure}[tb]
	\includegraphics[width=\linewidth]{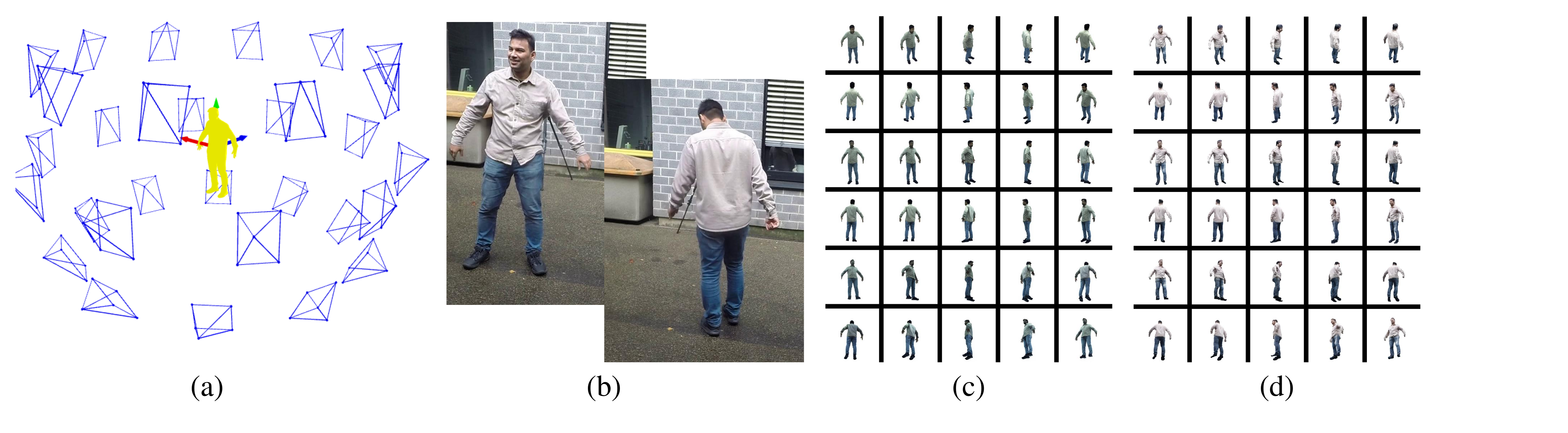}
	\caption
	{
	    Illustrations of image proxy.
            (a) Visualization of the camera distribution for rendering the proxy images. 
            (b) Monocular frames used to generate in-the-wild image proxy.
            (c) Proxy images in the dome.
            (d) Proxy images in the wild. 
            Best viewed with zoom.
	}
	\label{fig:image_proxy}
	%
\end{figure}
%
%
We propose the occlusion handling to address missing information when occlusion happens. To offer additional information for the occluded areas, we render proxy images of the human in canonical pose space. Depending on the lighting condition and camera setup, we have two configurations: in-the-dome and in-the-wild. Here, we demonstrate the details of proxy image generation (see Fig.~\ref{fig:image_proxy}).

\subsection{Proxy Image Generation in the Dome}
We have dense-view cameras for the in-the-dome case. Consequently, we select one frame to reconstruct its geometry and texture with space canonicalization. This enables us to render novel-view in-the-dome proxy images in the canonical space (see Fig.~\ref{fig:image_proxy} (c)).

\subsection{Proxy Image Generation in the Wild}
Under the in-the-wild scenarios, where sparse-view or monocular cameras are predominant, occlusion handling becomes particularly crucial, especially in monocular scenarios.  Consider the most challenging scenario, namely the monocular camera setup. In such situations, it is not feasible to directly reconstruct geometry and appearance like in-the-dome scenario. Instead, leveraging the capabilities of the meta prior, we fine-tune multiple frames (see Fig.~\ref{fig:image_proxy} (b)) into a unified canonical space simultaneously to construct a complete human representation. We then render it into novel-view in-the-wild proxy images in the canonical space (see Fig.~\ref{fig:image_proxy} (d)).

\section{Implementation Details} \label{sec:supp_implementation}
%
%
\begin{figure}[tb]
	\includegraphics[width=\linewidth]{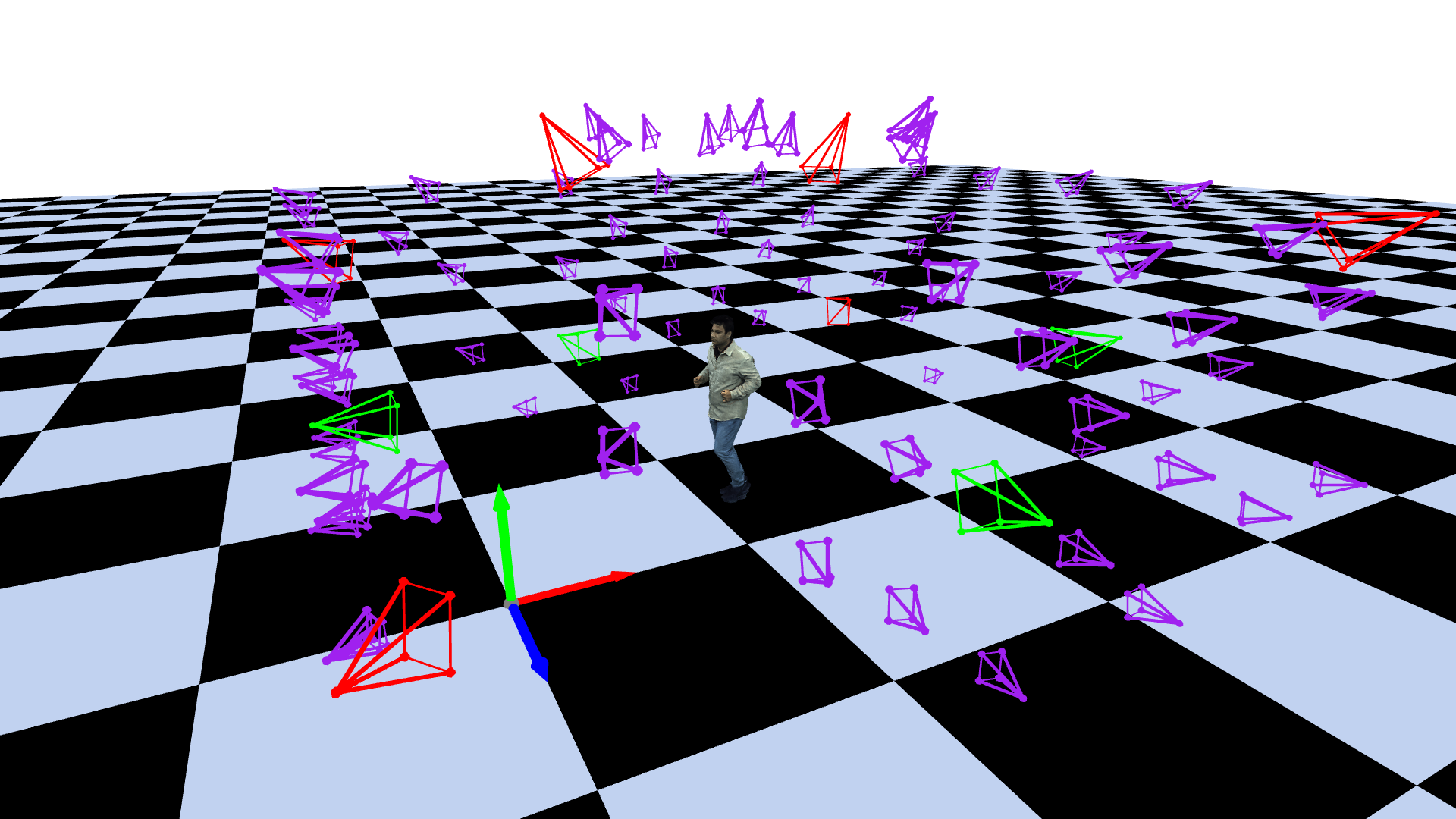}
	\caption
	{
	    Visualization of the camera distribution for prior learning, inference and evaluation in the comparison with state-of-the-art methods.
            Dense-view cameras with purple color are the training views used for prior learning.
            Four-view cameras with green color are the input views during inference.
            Six-view cameras with red color are the evaluation views.
	}
	\label{fig:camera_distri}
        \vspace{-10pt}
\end{figure}
%
%
In this section, we present implementation details of \model{} (Sec.~\ref{sub_sec:supp_metacap}), implementation details of methods that we compare with (Sec.~\ref{sub_sec:supp_comparison}), ablations (Sec.~\ref{sub_sec:supp_ablation}), and comparison on in-the-wild sequences (Sec.~\ref{sub_sec:supp_ITW}). Fig.~\ref{fig:camera_distri} illustrates the camera distribution when conducting the prior learning, fine-tuning and evaluation in the comparisons. These three sets of cameras do not overlap. Additionally, the motions used for prior learning and fine-tuning are distinct.

\subsection{\model{}} \label{sub_sec:supp_metacap}

\subsubsection{Space Canonicalization.} During comparison, we utilize the deformable template DDC~\cite{habermann2021} as the default template for space canonicalization. To acquire the template and the deformation parameters, we first create a character with smoothed template, embedded graph, skeleton and default motion. Subsequently, we follow the methodology outlined in \cite{habermann2021} to implement multi-view silhouette supervision using a differentiable renderer~\cite{diffrenderer} and distance transformation~\cite{fabbri20082d}. For the loose-cloth subject 'S5', we apply additional Chamfer loss supervision. For parametric models SMPL~\cite{SMPL:2015} and SMPL-X~\cite{SMPL-X:2019} used in the ablations and comparison methods, we first obtain 3D marker positions by animating the character's skeleton with the same motions used in the deformable template. We then utilize EasyMocap~\cite{ezmocap} to estimate the shape and pose parameters. 

\subsubsection{Meta-learning and Fine-tuning.}
During the meta-learning stage, we use approximately 100-frame images captured by 100-view cameras, paired with human template at each frame. We apply SGD~\cite{ruder2016overview} with $l^{out}=1.0$ to the outer loop and Adam~\cite{kingma2017adam} with $l^{in}=1e-4$ to the inner loop. 
In each outer loop sampling step, we randomly sample $M=24$ camera views and rays on each image simultaneously. After $M=24$ unrolled gradient steps in the inner loop, we follow Reptile~\cite{nichol2018first} to update the outer loop weights. 
The inner loop is warmed up by linearly updating learning rate from $1\%$ to $100\%$ for the first 50 outer loop steps.
The template threshold $\eta$ is set to 0.05 for `S2' and `S5', and to 0.01 for `S3' and `S27', with a threshold decay to $50\%$ applied after 300 outer loop steps. The total number of meta-learning outer loop steps is 2000. 
The input images are resized to $50\%$ and applied Gaussian blur with a $5*5$ kernel.
The weights of loss functions are set as $\lambda_\mathrm{c} = 10.0, \lambda_\mathrm{e} = 0.1, \lambda_\mathrm{m} = 0.1, \lambda_\mathrm{s} = 0.01$.

\par
During the fine-tuning stage, we load the meta-learned weights and apply the Adam optimizer~\cite{kingma2017adam} with learning rate $lr=1e-4$, $\beta_{1}=0.9$, $\beta_{1}=0.99$, $\epsilon=1e-15$ to fine-tune weights for 3000 steps. In each step, we randomly sample 8192 rays from all input observations. The template threshold $\eta$ is set 0.05. There's no warm-up in this stage. 
The weights of loss functions are set as $\lambda_\mathrm{c} = 10.0, \lambda_\mathrm{e} = 0.1, \lambda_\mathrm{m} = 0.1, \lambda_\mathrm{s} = 0.01$.

\subsection{Comparison Methods} \label{sub_sec:supp_comparison}
\subsubsection{DeepMultiCap.}
As DeepMultiCap~\cite{zheng2021deepmulticap} is trained on a large scale human scan dataset and exhibits generalization ability, we utilize the official checkpoint without additional fine-tuning. It relies on SMPL-X as the template model to provide geometry and global normal maps. Following the template procedure outlined earlier, we fit SMPL-X and subsequently render it to produce normal maps.

\subsubsection{DiffuStereo.}
Official DiffuStereo~\cite{shao2022diffustereo} utilizes  geometry results from DoubleField~\cite{shao2022doublefield} for initializing the disparity maps.
Since DoubleField~\cite{shao2022doublefield} is not open source, we employ the deformable template~\cite{habermann2021} as a substitute for initializing the disparity maps, as it contains rough geometry information. Subsequently, we use the official checkpoint trained with 20-degree angle images to refine the disparity maps. 
Due to the large camera baseline from 4-view cameras, the output point-clouds are often incomplete.
Therefore, we incorporate  additional template point-clouds to complete the mesh when applying the Possion surface reconstruction~\cite{kazhdan2006poisson}.

\subsubsection{Drivable Volumetric Avatars (DVA).}
The dataset division of 
DVA~\cite{remelli2022drivable} is the same as ours, including human template, training multi(dense)-view images and testing sparse-view images. We utilize the official code with our estimated SMPL-X parameters. 
We train the personalized DVA model using images from dense-view training set. At the testing stage, we adhere to the original paper's manner that no fine-tuning added, and employ sparse-view images and template to render novel view images.
Given that DVA does not focus on geometry reconstruction, we extract their estimated primitive parameters, convert them to box meshes, and use Possion surface reconstruction~\cite{kazhdan2006poisson} to reconstruct the final watertight mesh.

\subsubsection{TransHuman.}
TransHuman~\cite{Pan_2023_ICCV} is trained on multi-view videos with multiple subjects. However, We found that directly applying the official pre-trained checkpoint on our data yields low-quality results. Therefore, for each subject, we fine-tune them individually on our training set, and generate testing set results without additional fine-tuning.

\subsubsection{ARAH.}
ARAH~\cite{ARAH:ECCV:2022} incorporates a meta prior~\cite{MetaAvatar:NeurIPS:2021} trained from a large scale scan dataset. In our implementation, we use the official checkpoint as initialization and further fine-tune it with all the frames in the testing set. It's worth noting that other methods only utilize 1-frame sparse-view images as input during inference.

\subsection{Ablations} \label{sub_sec:supp_ablation}
\subsubsection{Weight Initialization and Space Canonicalization.}
During this ablation study, we maintain the camera setup consistent with the comparison section. Specifically, we utilize dense-view cameras for prior learning and four-view cameras for fine-tuning.
\par
We have three types of network initialization consisting of two baseline methods random weights, pre-trained weights and our meta weights.
Random weight initialization utilizes the default weight initialization from PyTorch~\cite{paszke2017automatic}.
Pre-trained weights are trained on the same views and frames as meta-learning. It's implemented by setting $M=1$ in meta-learning process and training for 1000 steps.
When performing the fine-tuning with random initialization and pre-trained initialization, we reserve 500 warm-up steps.
Meta weights are obtained following the procedure outlined in Sec.~\ref{sub_sec:supp_metacap}.
\par
In terms of space canonicalization, we employ three types: root canonicalization, SMPL canonicalization, and DDC canonicalization.
Root canonicalization is implemented by transforming world-space points to canonical space with the transformation of root joint of human from motions.
With SMPL template and SMPL motion parameters, we perform a more fine-grained canonicalization by determining the transformation of each world-space point to the nearest points.
When using DDC as the template, the canonicalization procedure is similar to SMPL template, but the transformation computation is adjusted for DDC.

\subsubsection{Number of Camera Views and Occlusion Handling.} In this ablation study, we investigate the impact of different camera views and occlusion handling. Specifically, we utilize DDC as the template for space canonicalization and meta weights for weight initialization.
\par
We first evaluate the effect of varying camera numbers in the meta-learning phase. We utilize 4-view cameras for fine-tuning, while we experiment with different camera numbers in meta prior learning: 1, 2, 4, 8, and dense.
\par
Next, we evaluate the influence of camera numbers in the fine-tuning phase. Here, we utilize dense-view cameras for prior-learning but different camera numbers in the fine-tuning: 1, 2, 4, 8. Additionally, we examine the influence of occlusion handling (OH) by employing this strategy during monocular fine-tuning.

\subsubsection{Convergence Speed and Quality}
We aim to investigate the the performance of convergence when using different weight initializations. The camera setup remains consistent with the comparison section and we utilize DDC as the template for space canonicalization. For dynamic evaluation during fine-tuning, we select a single frame from testing data. The corresponding qualitative results of the curves are presented in the supplementary video.

\subsection{Comparison on In-the-wild Sequences} \label{sub_sec:supp_ITW}
During the comparison on the in-the-wild sequences, we have four views to provide inputs and an additional view to offer ground truth images. We follow the Sec.~\ref{sub_sec:supp_metacap} and ~\ref{sub_sec:supp_comparison} to implement our method and ARAH.

\section{Learning and Testing Cost Against ARAH}
\label{sec:supp_cost}
We use 100 frames for the meta prior learning, it takes around 5-6 hours on a single GPU. 
When conducting fine-tuning, our method takes between 40 seconds and 3 minutes for one frame with one GPU. 
The prior used in ARAH is from MetaAvatar~\cite{MetaAvatar:NeurIPS:2021}, which uses 10-48 hours for the prior learning in a single GPU.
During fine-tuning, ARAH~\cite{ARAH:ECCV:2022} takes around 16 hours for 90 frames with 4 GPUs, i.e. around 10 minutes per frame. 
Thus, our method is significantly faster than ARAH during meta-learning and fine-tuning.

\section{Additional Comparisons on S3} \label{sec:supp_comparison_s3}
%
%
\begin{figure}[tb]
	\includegraphics[width=\linewidth]{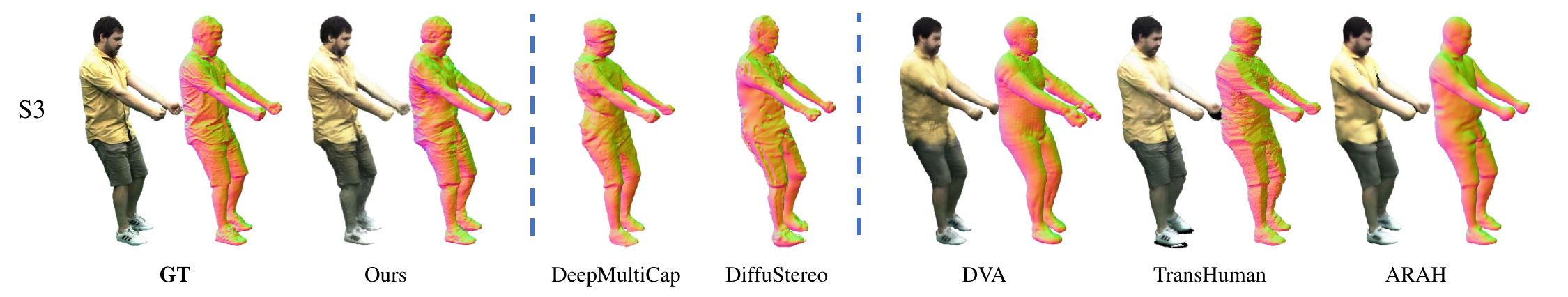}
	\caption
	{
        \textbf{Qualitative Comparison.}
	    We additionally compare our method with other approaches on S3 of DynaCap dataset.  Our method demonstrates superior performance in geometry capturing  and rendering quality.
	}
	\label{fig:comparison_sup}
\end{figure}
%
%
\begin{table}[t]
   \centering
       \caption{
  \textbf{Quantitative Comparison.}
  For the S3 from DynaCap dataset, our method still achieves state-of-the-art results for novel-view synthesis and geometry reconstruction.
  *Note, that ARAH requires 4D scans for meta learning and videos for the fine-tuning whereas other methods solely require static images.
  }
    \scalebox{0.75}{
    \begin{tabular}{|c|c|ccc|ccccc|}
    \hline 
    \multirow{2}{*} { \textbf{Method} } &  \multirow{2}{*} { \textbf{Subject} }    & \multicolumn{3}{c|}{\textbf{Appearance} } & \multicolumn{5}{c|}{ \textbf{Geometry} } \\
    \cline{3-10} & & \textbf{PSNR} $\uparrow$ & \textbf{SSIM} $\uparrow$ & \textbf{LPIPS} $\downarrow$  & \textbf{NC-Cos} $\downarrow$ & \textbf{NC-L2} $\downarrow$ & \textbf{Chamfer} $\downarrow$ & \textbf{P2S} $\downarrow$ & \textbf{IOU} $\uparrow$  \\

    \hline 
    \multirow{1}{*} { DeepMultiCap ~\cite{zheng2021deepmulticap} }
    & S3 & - & - & - & 0.131 & 0.425 & \bronze{1.137} & \bronze{1.158} & 0.717  \\
    
    \hline 
    \multirow{1}{*} { DiffStereo ~\cite{shao2022diffustereo}}
     & S3 & - & - & -  &  0.143 & 0.441 & 1.169 & 1.269 & \bronze{0.818} \\

    \hline 
    \multirow{1}{*} { DVA ~\cite{remelli2022drivable} }
    & S3 & 24.862 & 0.824 & 0.284  & \bronze{0.109} & \silve{0.378} & 1.593 & 2.119 & 0.465 \\

    \hline 
    \multirow{1}{*} { TransHuman ~\cite{Pan_2023_ICCV} } 
    & S3 & \silve{25.136} & \bronze{0.826} & \silve{0.277} & 0.118 & 0.393 & 1.477 & 2.006 & 0.797 \\ 

    \hline 
    \multirow{1}{*} { ARAH* ~\cite{ARAH:ECCV:2022}}
    & S3 & \bronze{25.093} & \gold{0.842} & \bronze{0.278}  & \gold{0.069} &  \gold{0.294} & \silve{0.780} & \silve{0.836} & \silve{0.866}  \\

    \hline 
    \multirow{1}{*} { \textbf{Ours} }
    & S3 & \gold{25.528} & \silve{0.839} & \gold{0.251} & \silve{0.106} & \bronze{0.382} & \gold{0.671} & \gold{0.792} & \gold{0.908} \\

    \hline
    \end{tabular}
    }
  \label{tab:quantitative_sup}	
\end{table}
Fig.~\ref{fig:comparison_sup} and Tab.~\ref{tab:quantitative_sup} present additional qualitative and quantitative results on the `S3' from DynaCap Dataset~\cite{habermann2021}. The implementations of the methods are consistent with those in the `Results' section. Our approach continues to outperform other methods in both rendering and reconstruction.

\section{Additional Comparisons on Monocular Methods} \label{sec:supp_comparison_s3_mono}
\begin{figure}[tb]
         \centering
	\includegraphics[width=0.95\linewidth]{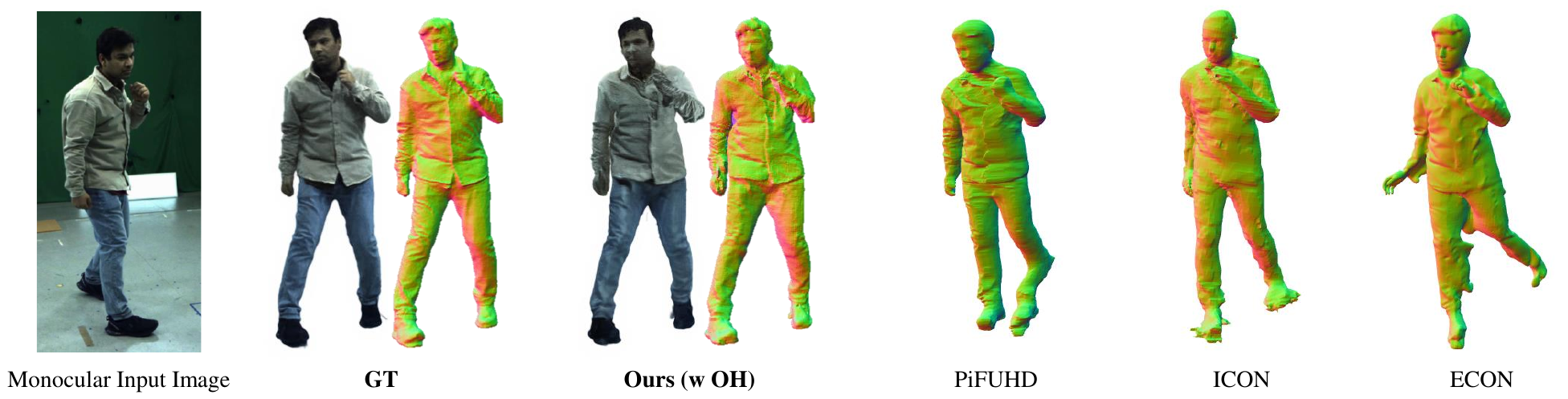}
        \caption
	{
        \textbf{Qualitative Comparison.}
        In this comparison, we compare our method, which involves monocular fine-tuning with occlusion handling, against other monocular reconstruction approaches, namely PiFUHD~\cite{saito2020pifuhd}, ICON~\cite{xiu2022icon}, and ECON~\cite{xiu2023econ}.
        Our method exhibits robustness to the human pose and camera pose, and produces superior geometry and appearance capture.
	}
	\label{fig:monocular}
\end{figure}
Fig.~\ref{fig:monocular} shows additional qualitative comparisons between  our method and monocular reconstruction methods. Here, we initialize our network with the meta prior and fine-tune it using monocular input images, with occlusion handling applied. Our approach employs perspective camera projection, enabling human reconstruction in real-world scale and coordinates. In contrast, PiFUHD~\cite{saito2020pifuhd}, ICON~\cite{xiu2022icon}, and ECON~\cite{xiu2023econ} utilize orthogonal camera projection.
PiFUHD~\cite{saito2020pifuhd} exhibits sensitivity to both human and camera poses. 
ICON~\cite{xiu2022icon} demonstrates limited generalization ability. 
ECON~\cite{xiu2023econ} predicts normal maps for the front and back sides, and integrates them onto SMPL template. The predicted normal maps may lack accuracy or fail easily.
Our method yields reasonable results by fine-tuning the canonical space human fields.
In the `Comparison' section, our method outperforms the multi-view method DeepMultiCap~\cite{zheng2021deepmulticap}, which presents superior results to multi-view PiFUHD.

\section{Ablation on Motion Capture Quality} \label{sec:supp_ablation_mocap}
\begin{table}[tb]
   \centering
     \caption{  \textbf{Quantitative Ablation.}
      Here, we study the influence of motion tracking quality on our method.
      Comparing to dense mocap, our method with sparse mocap exhibits a slight decrease in performance.}
    \scalebox{0.75}{
    \begin{tabular}{|c|c|c|ccc|ccccc|}
    \hline 
    \multirow{2}{*} { \textbf{Method} } & \multirow{2}{*} { \textbf{Motion} } &  \multirow{2}{*} { \textbf{Subject} }    & \multicolumn{3}{c|}{\textbf{Appearance} } & \multicolumn{5}{c|}{ \textbf{Geometry} } \\
    \cline{4-11} & & & \textbf{PSNR} $\uparrow$ & \textbf{SSIM} $\uparrow$ & \textbf{LPIPS} $\downarrow$  & \textbf{NC-Cos} $\downarrow$ & \textbf{NC-L2} $\downarrow$ & \textbf{Chamfer} $\downarrow$ & \textbf{P2S} $\downarrow$ & \textbf{IOU} $\uparrow$  \\

    \hline 
    ARAH*& Dense & S2 & \silve{26.279} & \bronze{0.833} & \bronze{0.302} & \gold{0.079} & \gold{0.315} & \bronze{0.839} & \bronze{0.913} & \bronze{0.859}  \\

    \hline 
    Ours & \multirow{1}{*} { Sparse}
    & S2 & \bronze{26.240} & \silve{0.836} & \silve{0.253} & \bronze{0.102} & \bronze{0.362} & \silve{0.712} & \silve{0.840} & \silve{0.883}  \\

    \hline 
    Ours& \multirow{1}{*} { Dense }
    & S2 & \gold{26.529} & \gold{0.841} & \gold{0.249} & \silve{0.096} & \silve{0.351} & \gold{0.679} & \gold{0.814} & \gold{0.887} \\

    \hline
    \end{tabular}
    }
  \label{tab:ablation_sparsemotion}	
\end{table}
To evaluate the influence of motion capture quality to our method, we replace motions from dense mocap with sparse mocap and generate rendering and reconstruction results. The sparse motions come from the same four-view camera setup used for fine-tuning, while the dense motions are estimated from 34 cameras in the dome. Tab.~\ref{tab:ablation_sparsemotion} demonstrates that, though the perfomance drops a bit, our method with sparse mocap still produces comparable rendering quality and better geometry compared to ARAH~\cite{ARAH:ECCV:2022} with dense mocap.

\section{Applications}\label{sec:supp_applications}
\subsection{Interpolation in Weight Space}
%
%
\begin{figure}[tb]
	\includegraphics[width=\linewidth]{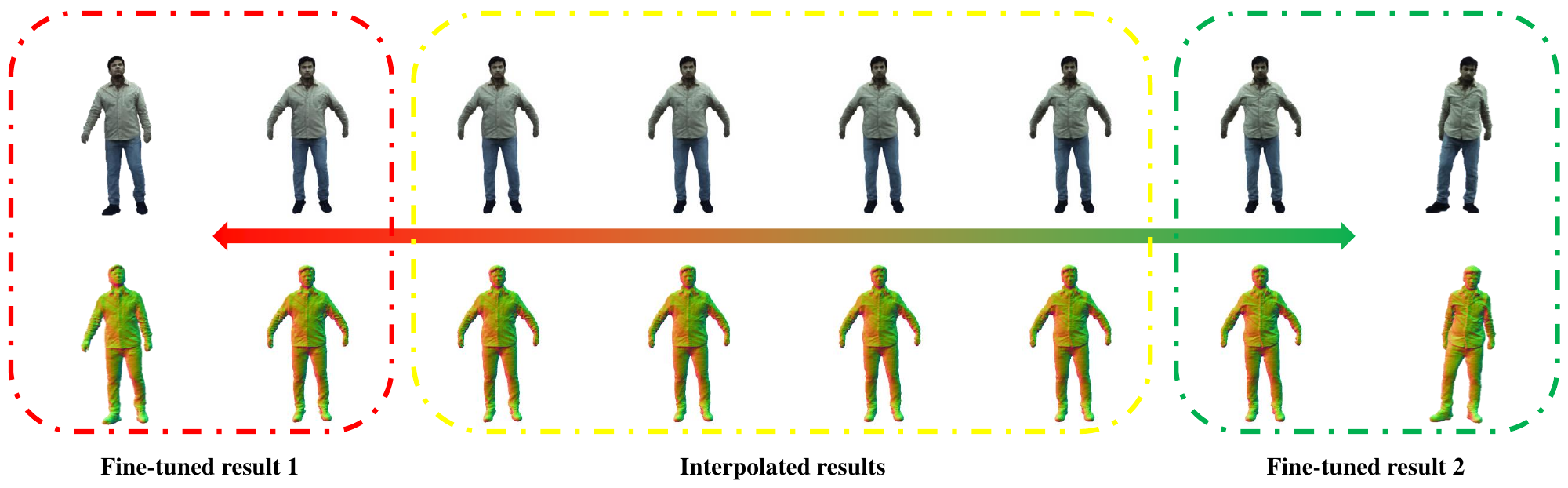}
	\caption
	{
	    Appearance and geometry interpolation on two fine-tuned results. The red and green boxes represent the appearance and geometry of different frames' fine-tuned results displayed in both world space and canonical space. 
	}
	\label{fig:interpolation}
\end{figure}
%
%
Thanks to the space canonicalization and meta initialization, we are able to linearly interpolate results from different frames in the weight (hyper) space and produce meaningful novel interpolated appearance and geometry results, as shown in Fig.~\ref{fig:interpolation}. This experiment further validates our hypothesis that space canonicalization narrows the range of spatial features and facilitates meta prior learning.

\subsection{Animating the Fine-tuned Results}
%
%
\begin{figure}[tb]
	\includegraphics[width=\linewidth]{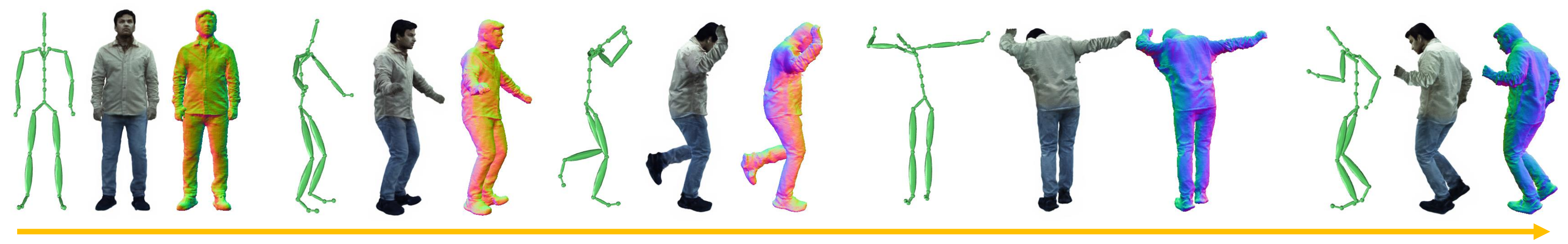}
	\caption
	{
	    Animating our four-view fine-tuned results over time. Our hybrid representation can be easily animated with motion and corresponding template.
	}
	\label{fig:animation}
\end{figure}
%
%
After fine-tuning our meta prior with four-view images, we obtain a canonicalized hybrid human avatar. This avatar can be easily animated with novel motions and corresponding deformable template, like Fig.~\ref{fig:animation}. The animated results maintain photorealistic appearance and high-quality geometry.

\section{Temporal Fine-tuned Results}\label{sec:supp_temporal}
\begin{figure}[H]
	\includegraphics[width=\linewidth]{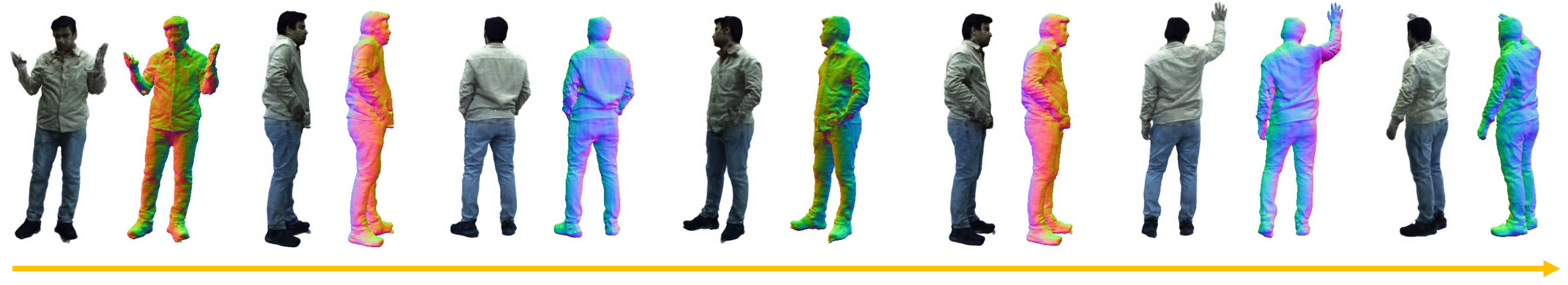}
	\caption
	{
	    Fine-tuned results on a temporal sequence using a frame-by-frame manner.
	}
	\label{fig:temporal}
\end{figure}

Fig.~\ref{fig:temporal} shows fine-tuned results on a temporal sequence. As our method is not designed for temporal inputs, we generate these results by a frame-by-frame fine-tuning. Please refer to the project page for the video display.
%
%
%
%
\end{document}